\documentclass[preprint,12pt]{elsarticle}
\usepackage[english]{babel}
\usepackage{amsfonts}
\usepackage[ruled,vlined]{algorithm2e}
\usepackage{textcomp}
\usepackage{csquotes}
\usepackage{commath}
\usepackage{xspace}
\usepackage{array}
\usepackage{tabularx}
\usepackage{nicefrac}
\usepackage{harpoon}
\usepackage{bm}
\usepackage{color}
\usepackage[pdftex,dvipsnames]{xcolor}
\usepackage{multirow}
\usepackage{booktabs}
\usepackage{hyperref}
\usepackage{cleveref}
\crefname{table}{table}{tables}
\Crefname{table}{Table}{Tables}
\crefname{figure}{figure}{figures}
\Crefname{figure}{Figure}{Figures}
\crefname{equation}{equation}{equations}
\Crefname{equation}{Equation}{Equations}
\crefname{section}{}{}
\Crefname{section}{}{}
\usepackage{subcaption}
\usepackage{graphicx}
\usepackage{rotating}
\usepackage{tabularx}
\usepackage{url}
\date{23 May 2025}

\usepackage{pifont}

\usepackage[colorinlistoftodos,prependcaption, textsize=normalsize]{todonotes}

\let\oldbibliography\thebibliography
\renewcommand{\thebibliography}[1]{%
  \oldbibliography{#1}%
  \setlength{\itemsep}{0pt}
  \setlength{\parsep}{0pt}
}

\begin{document}

\begin{titlepage}
    \centering
    \vspace*{1cm} 
    
    \Large\textbf{Accepted Manuscript} \\
    \vspace{1cm}
    
    \LARGE \textbf{Real-Time Frame- and Event-based Object Detection with Spiking Neural Networks on Edge Neuromorphic Hardware: Design, Deployment and Benchmark} \\
    \vspace{1cm}
    
    \large \textbf{Authors:} \\
    Udayanga G.W.K.N. Gamage, Yan Zeng, Cesar Cadena, Matteo Fumagalli, Silvia Tolu \\
    \vspace{1.5cm}
    
    \normalsize
\textbf{Published in:} \textit{Neurocomputing} (Elsevier), 2026 \\
{\large\textbf{DOI:} \href{https://doi.org/10.1016/j.neucom.2026.133820}{10.1016/j.neucom.2026.133820}\par}
\vspace{1.5cm}
    
    \fbox{
\parbox{0.8\textwidth}{
\centering
\small
© 2026. This is the author’s accepted manuscript of an article published in\\
\textit{Neurocomputing (Elsevier)}.\\
The final published version is available at the DOI above.\\
This manuscript is made available under the \textbf{CC-BY-NC-ND 4.0} license.
}
}
    
    \vfill
\end{titlepage}
\setcounter{page}{1} 

\begin{frontmatter}
\title{Real-Time Frame- and Event-based Object Detection with Spiking Neural Networks on Edge Neuromorphic Hardware: Design, Deployment and Benchmark}

\author[DTU]{Udayanga G.W.K.N. Gamage\corref{cor1}}
\author[DTU]{Yan Zeng}
\author[ETH]{Cesar Cadena}
\author[DTU]{Matteo Fumagalli}
\author[DTU]{Silvia Tolu}

\cortext[cor1]{Corresponding author:\\
Email: udayangakn@eie.ruh.ac.lk\\
Present address: Faculty of Engineering, University of Ruhuna, Hapugala,Galle 80000, Sri Lanka}

\affiliation[DTU]{organization={Department of Electrical and Photonics Engineering, Technical University of Denmark},
            addressline={Anker Engelunds Vej 1},
            city={2800 Kongens Lyngby},
            country={Denmark}}
            
\affiliation[ETH]{organization={Autonomous Systems Lab, ETH zürich},
            addressline={Leonhardstrasse 21},
            city={8092 Zürich},
            country={Switzerland}}

\begin{abstract}

Real-time object detection on energy-constrained platforms is critical for applications such as UAV-based inspection, autonomous navigation, and mobile robotics. Spiking neural networks (SNNs) on neuromorphic hardware are believed to be significantly more energy-efficient than conventional artificial neural networks (ANNs). In this work, we present a comprehensive methodology for designing general SNN detection architectures targeting neuromorphic platforms, along with the engineering adaptations required to deploy them on the state-of-the-art Neuromorphic processor, Intel Loihi 2.

We developed three SNN-based object detection models and benchmarked on Loihi 2 using both frame-based and event-based datasets, comparing performance with ANN-based detection on the NVIDIA Jetson Orin Nano, NVIDIA Jetson Nano B01, and the Apple M2 CPU. Our results show that SNNs on Loihi 2 can perform real-time detection while achieving the lowest per-inference dynamic energy among all platforms. Also, Loihi 2 outperforms the other platforms in terms of power consumption, though ANNs on Jetson Orin Nano achieve higher inference rates.

\textcolor{black}{Furthermore, our ANN-to-SNN distillation-aware SNN direct training approach enables SNNs to recover 87–100\% of the detection accuracy of their ANN counterparts while maintaining lower inference latency than conventional ANN-to-SNN conversion methods.These results highlight the potential of neuromorphic systems for energy-efficient, real-time object detection at the edge.}

\end{abstract}


\end{frontmatter}

\section{Introduction}
Object detection, which involves the identification and localization of objects in visual scenes, is a fundamental capability for intelligent perception systems. Applications such as autonomous driving, drones, mobile robots, and smart surveillance rely on real-time object detection to perceive their surroundings accurately and act safely \cite{autonomous_uav},\cite{autonomous_drive}. For instance, a self-driving car must immediately detect pedestrians, vehicles, and traffic signs to ensure safe navigation, while a drone must continuously identify obstacles to avoid collisions.
Beyond navigation, real-time object detection is critical for inspection activities, particularly in civil infrastructure monitoring. The ability to detect and respond to anomalies such as cracks or structural deformations immediately on-site can reduce maintenance costs, minimize downtime, and prevent catastrophic failures, thus ensuring public safety \cite{structural_inspection_review}.

In the literature, frame-based object detection using grayscale or RGB camera images is widely used.
Under extremely low lighting conditions, these frame-based cameras produce noisy, blurry, and low-contrast images. In extremely bright conditions, scenes become overexposed, causing saturated highlights, loss of texture, glare, and blooming \cite{low_light_img_review}. Additionally, frame-based cameras cannot simultaneously capture dark and bright regions in high dynamic range scenes \cite{event_vision_survey}.
In contrast, thanks to their high dynamic range, Dynamic Vision Sensors (DVS) also known as event-based cameras, can capture scenes across a wide range of lighting conditions including extremely low light and overexposed areas, while reducing noise, motion blur, and saturation, resulting in objects that appear sharper and clearer than in frame-based images \cite{event_vision_survey}.

For frame-based and event-based real-time object detection, deep artificial neural network (ANN) detectors such as YOLO \cite{YOLOv3_ref} and SSD \cite{SSD_ref} have proven promising, outperforming model-based or classical machine learning methods in terms of detection accuracy.
Many of these standard deep object detectors, although highly accurate, have a large number of model parameters, making them large in size and requiring high-end computing and memory resources to run in real-time \cite{resource_limit_review}. Consequently, in scenarios where these standard models are used, captured frames or events are sent to remote servers that provide the necessary computational and memory resources.
However, relying on remote processing introduces communication latency and creates a dependency on stable network connectivity. These factors significantly hinder real-time responsiveness, particularly in remote sensing applications or bandwidth-constrained environments.

In contrast, performing edge device detection, i.e., directly on platforms such as drones, mobile devices, or embedded processors, it eliminates communication delays and ensures real-time operation. This edge-based approach enhances reliability, safety, and autonomy, even in situations where network connectivity is intermittent, limited, or unavailable \cite{lightweight_edge_devices}.
In addition to avoiding communication delays and network issues,  edge device detection improves  energy efficiency, since fewer data need to be transmitted, and enhances scalability, since each device processes its own data instead of relying on a central server \cite{lightweight_edge_devices}.

To run deep learning models on the edge, energy-efficient and lightweight processing units such as NVIDIA Jetson, Google Coral, and Intel Movidius have been developed \cite{edge_device_comparison}. Energy efficiency is particularly critical for battery-powered platforms such as UAVs and mobile robots, as it directly affects mission duration and area coverage, while still maintaining real-time performance without compromising detection accuracy, a challenging balance to achieve. However, compared to traditional high-end GPU platforms, these edge processors have limited resources, such as lower computing power and memory. Consequently, standard deep object detectors cannot run in real-time on these devices.
To address this, the literature has proposed lightweight real-time ANN-based detection models, such as EdgeYOLO and SSD with a MobileNet backbone, specifically designed for deployment on resource-constrained edge platforms \cite{edgeYOLO2023},\cite{SSD_MobileNet2025}.

In addition to traditional von Neumann load-store architectures (e.g., Jetson), neuromorphic processors have recently been successfully deployed at the edge for real-time operations, such as obstacle avoidance in drone navigation \cite{neuro_drone_journal},\cite{neuro_drone_obstacle}. Deep learning architectures can also be run on these neuromorphic platforms, but not as conventional ANNs, instead, they are implemented as spiking neural networks (SNNs). Compared to ANNs, SNNs on neuromorphic hardware are considered to be more energy-efficient \cite{Roy2019}. 

Despite their theoretical promise, SNNs have been scarcely explored on neuromorphic hardware for high-stakes inspection tasks, and their energy efficiency, latency, and real-time performance remain poorly understood. Existing studies, whether on GPUs/CPUs or using ANNs, primarily focus on regular objects, leaving the detection of irregular or anomalous objects in event-driven data largely unaddressed \cite{event_vision_survey}.

\textcolor{black}{We address these gaps through two key contributions. First, we propose a comprehensive methodology for designing general SNN-based object detection architectures tailored to neuromorphic platforms, along with the necessary engineering adaptations for deployment on the state-of-the-art Intel Loihi 2.
Second, we present a systematic benchmark of SNN-based object detection on Loihi 2 using both frame-based and event-based datasets, and compare its performance against ANN-based detection on the NVIDIA Jetson Orin Nano, NVIDIA Jetson Nano B01, and the Apple M2 CPU. Our evaluation encompasses both regular, well-defined objects and irregularly shaped structural anomalies, thereby reflecting a broad range of object detection scenarios.}

The main contributions of this paper are as follows:

\begin{itemize}
    \item Development of three lightweight SNN architectures for Intel Loihi 2 and a complete pipeline for implementation, training, deployment, and benchmarking on neuromorphic hardware.
    
    \item Adaptation an ANN-to-SNN knowledge distillation framework to enhance the detection accuracy of directly trained SNNs.
    
    \item Cross-platform benchmarking of SNNs on Loihi 2 versus ANN counterparts on Jetson Nano (edge GPU) and MacBook M2, evaluating detection performance, inference rate, energy consumption, and energy–delay product (EDP).
    
    \item \textcolor{black}{Introduction of a UAV-based grayscale tunnel inspection video dataset for defect detection, with cross-platform benchmarking and enabling generalization of original benchmarks to field experiments}
\end{itemize}

\section{\textbf{Background}}

In this section, we provide fundamental knowledge on Dynamic Vision Sensors (DVS) (event-based cameras) and discuss how their data can be encoded for input into deep learning models for object detection and classification. We then introduce the basics of SNNs and provide an overview of neuromorphic processors. Finally, we cover essential concepts of quantization and knowledge distillation.

\subsection{Dynamic Vision Sensors(DVS)}

DVS cameras operate asynchronously, generating events whenever the brightness at a pixel changes beyond a predefined threshold. Positive events indicate an increase in brightness, while negative events indicate a decrease \cite{v2e_paper}. By capturing only changes rather than full frames, DVS offer several advantages over traditional cameras, including ultra-low latency, minimal motion blur, high dynamic range, and low power consumption \cite{event_vision_survey}.

For downstream tasks such as classification or object detection, these raw events are typically converted into structured tensor representations. Common encoding methods include 2D 2-channel event histograms \cite{hats}, voxel-grids \cite{voxelgrid}, and event cubes \cite{LOIC}, which organize the sparse event data into formats suitable for deep learning models.
\textcolor{black}{In this work, we evaluate two different event-based datasets for multi-object detection and localization tasks. For event-based data encoding, we employ 2D two-channel event histograms and voxel grid representations, as these approaches are widely used in object detection literature, including the works discussed in the Related work section.}

\subsection{Spiking Neural Networks}

Spiking Neural Networks (SNNs) process information using discrete spike events inspired by biological neural communication \cite{snn_survey}. Their sparse, event-driven operation enables high energy efficiency and low power consumption, making them well suited for neuromorphic hardware and real-time applications. Unlike conventional neural networks with continuous activations, SNNs employ biologically inspired neuron models such as the leaky integrate-and-fire (LIF) neuron \cite{pLIF}.

In SNNs, neurons such as LIF integrate incoming synaptic currents over time and emit a spike when the membrane potential exceeds a firing threshold, producing a binary output. Network inference is performed over a fixed number of discrete time steps.
Each neuron emits at most one spike per time step, after which the membrane potential is either reset or partially retained \cite{pLIF}.
\textcolor{black}{We adopt SNNs with LIF neurons, as these neurons are natively supported by the Intel Loihi 2 chip on which our models are deployed to benchmark.}

SNNs can be trained using offline or online strategies, with offline approaches being more common in deep learning. These include ANN-to-SNN conversion \cite{annsnnconversion} and direct training using algorithms such as backpropagation through time (BPTT) \cite{snn_backprop} or Slayer \cite{slayer}.
Although ANN-to-SNN conversion often yields higher accuracy, it generally results in higher inference latency than direct SNN training due to the need for many time steps \cite{panda2021snn}.
\textcolor{black}{So in our work, we used SNN direct training to reduce time steps required for an inference inturn it reduces the nference latency.}

\subsection{Neuromorphic Processors and Intel Loihi 2}

Neuromorphic processors are specialized hardware designed to emulate brain-like computation using arrays of artificial neurons and synapses that communicate through sparse, event-driven spikes \cite{ibm_neuromorphic_computing}. This architecture enables highly parallel, low-latency, and energy-efficient processing, making neuromorphic hardware well suited for tasks such as sensory processing and pattern recognition. Notable neuromorphic platforms include Akida \cite{brainchip_akida}, Dynapse \cite{synsense_dynap_cnn}, SpiNNaker \cite{furber2014spinnaker}, Intel Loihi \cite{intel_loihi}, and Loihi 2 \cite{intel_loihi2}.

Intel Loihi 2 integrates over one million hardware-fabricated neurons, such as leaky integrate-and-fire (LIF) neurons, enabling fully event-driven computation. As a result, traditional ANN models cannot be executed directly on the chip and must instead be expressed as spiking neural network (SNN) models.
Development for Loihi 2 is supported by Intel’s open-source LAVA framework \cite{lava_framework_video}, including lava-dl \cite{lava_dl}, which allows model simulation on CPUs and deployment on neuromorphic hardware. While evaluation of performance and energy efficiency requires execution on physical Loihi 2 chips, remote access is available through Intel’s Oheo Gulch cluster \cite{loihi2_oheo_gulch}. Loihi 2 is not yet commercially available, and several hardware features remain under development.
\textcolor{black}{To evaluate our SNN models on a neuromorphic platform, we used the Intel Loihi 2 chip on the Oheo Gulch development board, accessed remotely.}

\subsection{Quantization}

Quantization reduces the numerical precision of weights and activations typically from floating-point (FP32/FP16) to lower-bit integers (e.g., INT8) \cite{mit_quantization_lecture2020}, decreasing model size, memory usage, and computation. While it can affect accuracy, techniques like quantization-aware training (QAT) help models compensate for precision loss, enabling fast and energy-efficient inference on edge devices.
\textcolor{black}{In this work, we employ quantization-aware training (QAT) for SNNs using 8-bit weight quantization.}

\subsection{Knowledge Distillation}

Knowledge distillation trains a smaller student model to mimic a larger teacher model by learning teacher model's soft outputs rather than just hard labels \cite{mit2020knowledge}. This approach improves generalization while reducing computation and memory. Distillation can transfer outputs, intermediate features, gradients, or attention maps using task-specific loss functions, enabling the student to efficiently learn both semantic and structural behaviors of the teacher, as in object detection \cite{mit2020knowledge}.
\textcolor{black}{In this work, we combine ANN-to-SNN knowledge distillation with direct SNN training to improve detection accuracy of SNNs.}

\section{Related Work}

Recent SNN research for DVS data has progressed from simple classification to complex vision tasks, yet achieving efficient multi-object detection on neuromorphic hardware remains an open challenge.

\textcolor{black}{\subsubsection*{Spiking Neuron Dynamics and Constraints:}}
While biologically plausible models like LIF neurons with hard-reset mechanisms improve temporal feature extraction over non-leaky models \cite{VicenteSola2025}, many existing designs remain hardware-incompatible. For instance, infinite-threshold accumulators \cite{Barchid2022} or models relying on simulated event streams \cite{iwann23} fail to address the finite-precision constraints and "reality gap" of physical neuromorphic processors like Loihi 2.

\textcolor{black}{\subsubsection*{SNN-based Object Detection:}}
Existing SNN-based object detection paradigms generally follow two main trajectories, both of which face significant deployment hurdles on neuromorphic hardware. Conversion-based methods, while effective at preserving ANN-level precision \cite{SpikingYOLO}, typically incur prohibitive inference latency due to the high number of timesteps required for rate-coded convergence. Conversely, directly trained SNN detectors—such as those based on SpikeYOLO \cite{SpikeYOLO} or SSD-style architectures \cite{LOIC}—frequently exceed the rigid neuron and synapse budgets of a single Loihi 2 chip. Notably, even parameter-efficient backbones like MobileNetv2 \cite{SSD_MobileNet2025} prove unsuitable for fully on-chip deployment, as their depthwise-separable designs produce large activation volumes that inflate per-core memory demand. These challenges are further exacerbated in hybrid ANN–SNN models \cite{hybrid_ANN_SNN1} and recurrent spiking networks \cite{Recur_SNN}, where memory constraints often necessitate off-chip communication, thereby degrading the inherent energy efficiency of the neuromorphic processor.

\textcolor{black}{\subsubsection*{Neuromorphic hardware targeted Deployments:}}

Most deep SNNs on Loihi 2 are limited to classification or segmentation (e.g., car classification \cite{CarSNN}, lane segmentation \cite{LaneSNN}). While efficient, these classification-centric models do not address the architectural complexity of bounding box regression. Recent attempts at detection using Sigma-Delta Neural Networks (SDNNs) \cite{SDNN_Loihi2} achieved 314 FPS but differ fundamentally from conventional SNNs by transmitting real-valued differences rather than binary spikes, leaving their performance on sparse, asynchronous event data unclear.

\textcolor{black}{\subsubsection*{ANN-based object detection at the edge:}}

While ANN-based platforms such as NVIDIA Jetson remain dominant for edge object detection, ultra-low-power microcontrollers like GAP9 are emerging. For example, a YOLOv8-based model on GAP9 achieved 30\% mAP@0.5:0.5 on Pascal VOC with 16.9 ms latency, 94.1 mW power, and 1.59 mJ energy per inference \cite{gap9_tinyyolo}.
These results highlight the performance-efficiency trade-offs across edge platforms and underscore the critical absence of hardware-measured, fully spiking multi-object detection on Loihi 2.

\textcolor{black}{Based on the limitations identified in prior work, a key research gap emerges: the lack of studies on SNNs that are deployable on neuromorphic processors for both frame-based and event-based object detection. Additionally, there is a scarcity of on-chip benchmarking of frame-based and event-based SNNs for multi-object detection using fully spiking models. Consequently, real-world evaluations of latency, throughput, and energy efficiency on modern neuromorphic platforms such as Loihi 2 remain largely unexplored.}

\section{\textbf{Methods}}

\begin{figure*}[!t]
\centering
\includegraphics[width=1.0\textwidth]{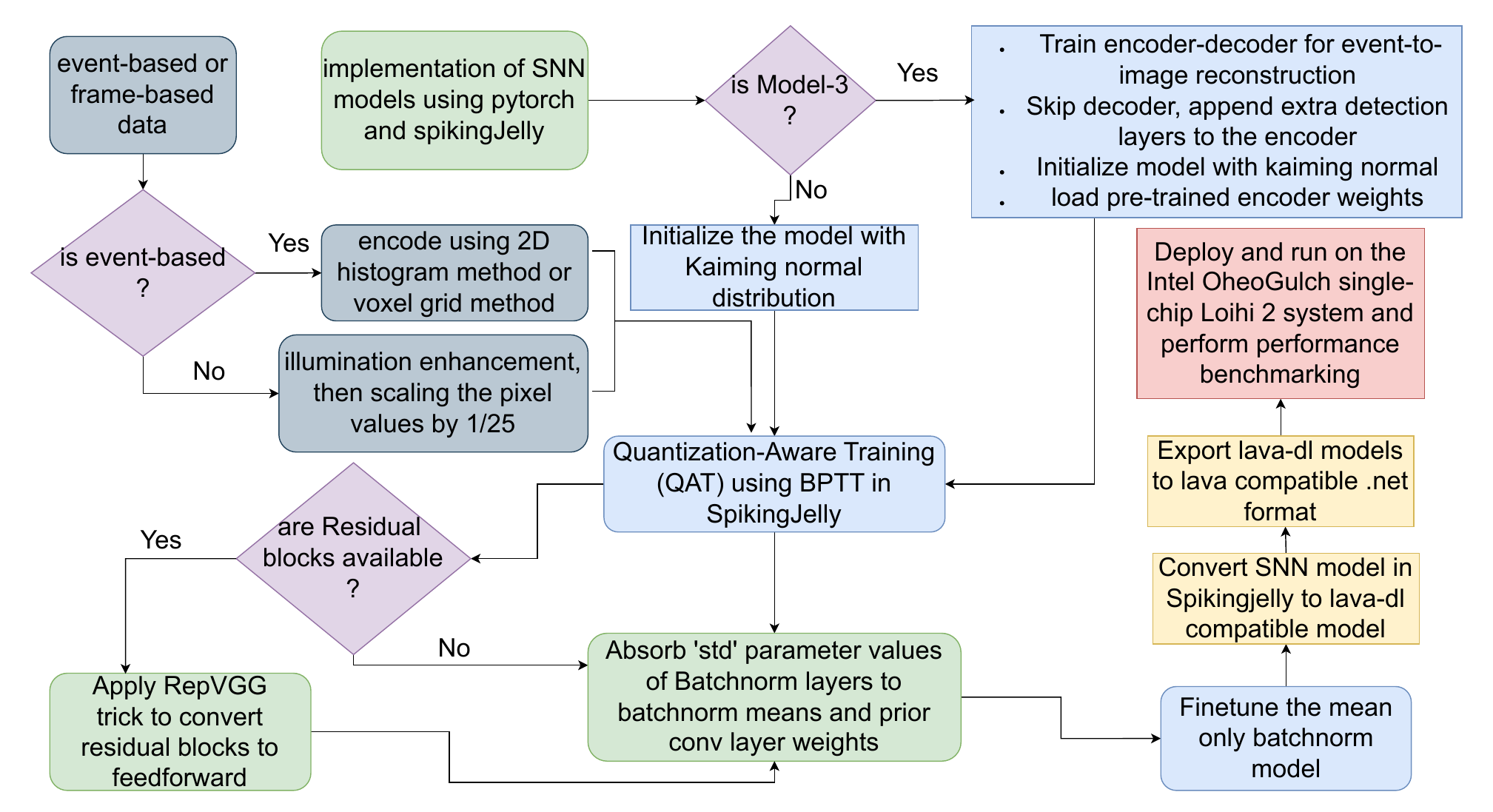}%
\caption{Overall workflow for SNN model development, training, deployment on Loihi 2, and benchmark.}
\label{fig:overall_setup}
\end{figure*}

In this section, we describe the methods used in the overall workflow illustrated in \cref{fig:overall_setup}. These include the development methodology for the SNN and ANN object-detection models, the event-based data encoding approach, the detection and distillation losses employed during training, and the evaluation metrics used for benchmarking the models.

\textcolor{black}{\subsection{SNN Model Development}}

We developed three models, each incorporating a distinct backbone architecture and a unified anchor-free coupled detection head.
All models were designed to contain fewer than one million neurons to ensure that they could be deployed on a single Loihi 2 chip. During model development, we adhered to several SNN-specific design constraints, such as the removal of bias terms and the exclusion of max-pooling layers \cite{snn_constraints}.
In all our proposed models, max-pooling layers were replaced with strided convolution layers (stride = 2) placed after each convolutional layer, thereby reducing the spatial resolution of the feature maps without relying on non-spiking-compatible pooling operations.
\textcolor{black}{Additional constraints arose from ensuring compatibility with neuromorphic hardware.
For example, each model is required to have a fixed number of neurons, which prevents inference on dynamically varying rectangular spatial input dimensions, a capability that is common on conventional microprocessors \cite{neuromporphic_constraints}.
Hence for SNNs on Loihi 2, they require distinct compilations for target hardware for every resolution change and hence cannot perform  multi-scale detection for different input resolutions.}

We also accounted for the current limitations specific to Loihi 2, as it remains a research-grade neuromorphic platform \cite{inrc_deep_learning_loihi2}.
One such limitation is the lack of support for branching. Consequently, we had to avoid using multiple branches, which would have allowed multiple heads for detection at different scales \cite{inrc_deep_learning_loihi2}.
For the same reason, rather than using decoupled heads where classification and regression tasks are handled by separate feature-specific heads, we had to use a single, coupled head.

\textcolor{black}{The latest lightweight YOLO models focus on reducing the number of parameters by leveraging inverted bottleneck architectures, such as pointwise and depthwise convolutions. While these approaches decrease parameter count, they drastically increase the number of activation neurons. For example, although SOTA lightweight edge-focused YOLOv8n \cite{yolov8_ref} ANN has only about 3 million parameters with a $3 \times 320 \times 320$ input resolution, it generates around 16 million activation neurons.  
On Loihi 2, activation neurons are pre-determined and pre-allocated as hardware units based on the model architecture and input resolution. 
Therefore, when designing SNN models, it is more important to reduce the number of activation neurons rather than just the model parameters. Accordingly, for our lightweight models, we use conventional 2D convolutional layers—where each kernel operates across all input channels—rather than edge-aware architectures based on pointwise or depthwise convolutions \cite{SSD_MobileNet2025}.}

In all three models, the basic architectural block consists of a Conv2D layer followed by BatchNorm2D and LIF neurons with hard reset. 
Although Loihi 2 requires mean-only batch normalization for deployment, we designed SNNs using standard BatchNorm2D (mean and variance) during training to preserve expressiveness and stability. After training, we absorb the BN’s standard deviation into the preceding convolution weights and retain only the mean, yielding a mean-only BN that preserves the learned activation scaling with minimal loss of expressiveness.

We employ a hard-reset mechanism within the LIF dynamics, aligning with the findings in \cite{VicenteSola2025} because such an approach prevents voltage stagnation and enhances the extraction of temporal features from  event-based data.
In \cref{fig:model_architectures}, each “Conv2D” block in the backbone networks represents this sequence: Conv2D $\rightarrow$ BatchNorm2D $\rightarrow$ LIF.

The output layer, however, omits BatchNorm2D and consists solely of a Conv2D layer followed by LIF neurons with a hard reset. While LIF neurons in the preceding layers use a standard threshold to allow spiking, the output-layer LIF neurons are assigned a very high voltage threshold to prevent spiking. This design allows the membrane potentials of the output neurons to accumulate over the inference timesteps, enabling the final membrane potential to be interpreted directly as an analog output which is an essential requirement for bounding box regression in object detection.

All residual blocks in the models were implemented as RepVGG \cite{repvgg} blocks, allowing the residual connections to be reparameterized into equivalent feedforward connections at inference time, enabling faster inference.

As Intel Loihi 2 does not currently support branching in convolutional neural networks (i.e., splitting feature maps into multiple parallel heads), we are unable to employ multiple detection heads for handling different object scales. Consequently, all models in this work utilize a single detection head.
To address this limitation, we adopt an anchor-free detection head, as it is better suited for handling objects across a wide range of scales and aspect ratios compared anchor-based counterparts which requires manually defined anchor boxes \cite{anchorfree_pros}.
Our empirical evaluations further confirm that anchor-free variants consistently outperform their anchor-based counterparts across all evaluated datasets.
In standard object detection architectures, an anchor-free detection head is typically decoupled into two parallel branches: one for classification logits and another for bounding box regression. However, due to the same restriction on CNN branching in Loihi 2, we instead employ a coupled head design, where classification and regression are jointly predicted within a unified head.

\subsubsection{Backbone networks}

\paragraph{Model-1 backbone}

\begin{figure*}[!t]
\centering
\includegraphics[width=1.0\textwidth]{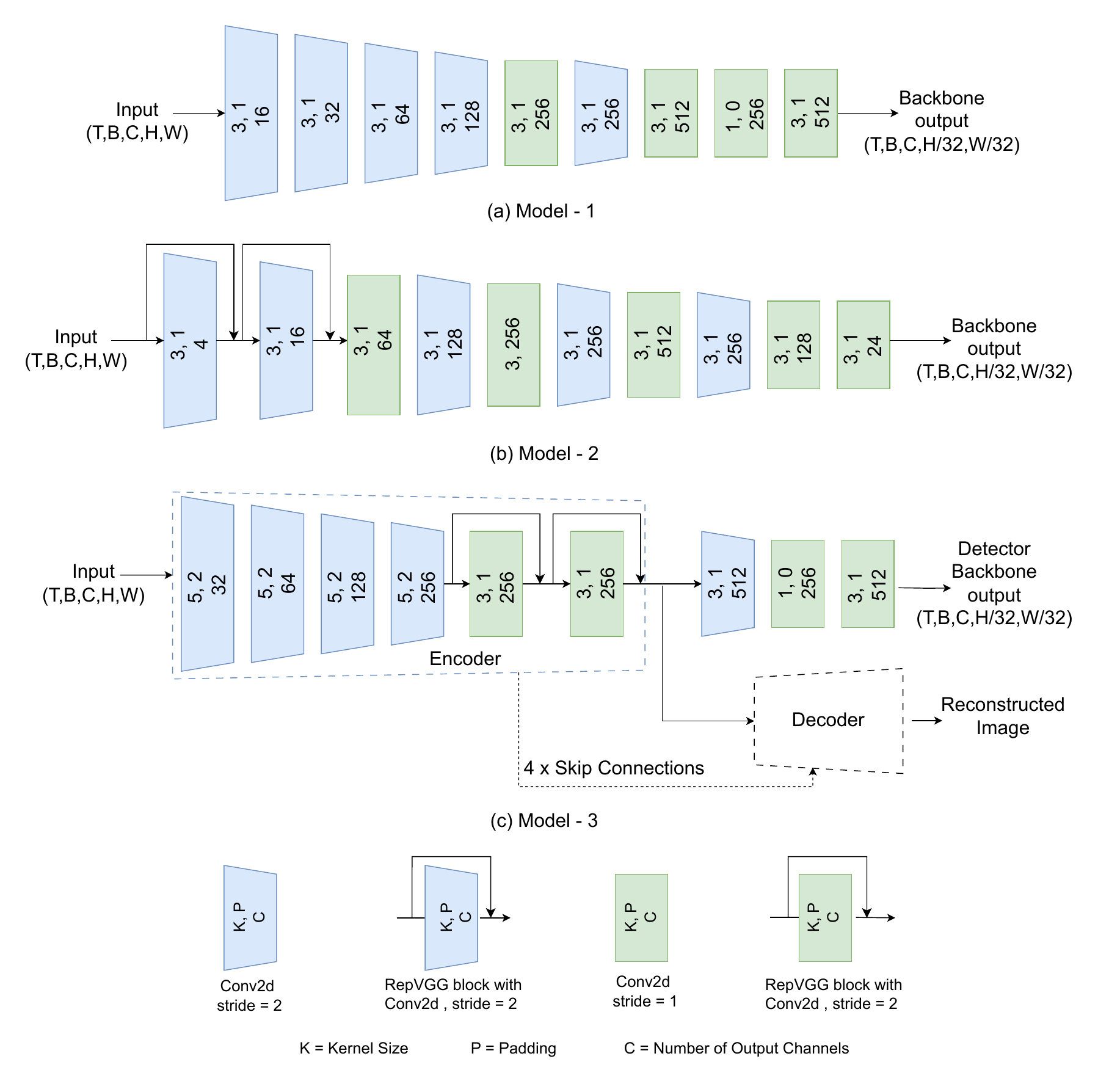}%
\caption{Proposed SNN model backbones for (a) Model-1 (b) Model-2 (c) Model-3. Each Conv2D layer of every model is followed by BatchNorm2D layer followed by LIF neurons as the non-linear activations.}
\label{fig:model_architectures}
\end{figure*}

The backbone of Model-1 was obtained by modifying the YOLOv3-tiny \cite{YOLOv3_ref} architecture. As shown in \cref{fig:model_architectures}.a, the backbone structure is illustrated in detail.

\paragraph{Model-2 backbone}

The backbone of Model-2 was inspired by the TY-v1.3-Small architecture proposed in \cite{gap9_tinyyolo} and modified as shown in \cref{fig:model_architectures}.b. The first two layers consist of residual blocks with a reduced number of channels, substantially decreasing the overall neuron count of the model. The residual connections in these early layers help mitigate any potential loss of representational capacity caused by the channel reduction, as they preserve and forward input information to deeper stages of the network.

From the third layer onward, the general architectural pattern of the TY-v1.3-Small design is retained. In these deeper layers, the increased number of feature maps enables the network to extract more complex representations, effectively leveraging the information preserved by the initial residual blocks.

\paragraph{Model-3 backbone}

The backbone of Model-3 was developed by leveraging the encoder of a modified event-to-image reconstructor, allowing us to reuse feature representations originally learned within an encoder–decoder reconstruction framework.
To implement this, we first adapted the architecture of the event-to-image reconstructor proposed in \cite{rebecq}, applying the following modifications.

In the original reconstructor, batch normalization layers were intentionally omitted to preserve fine-grained temporal and spatial details essential for high-fidelity image reconstruction. However, our objective is object detection, where stable feature representations are required across variations in object appearance and event statistics. In this context, batch normalization improves training stability, accelerates convergence, and enhances generalization \cite{batchnorm}. Therefore, we incorporated batch normalization after every convolutional layer in the encoder–decoder structure.

We further modified the architecture by removing the Conv-LSTM blocks, making the model fully convolutional and thus compatible with deployment on Loihi 2.

The final Model-3 backbone consists of the reused encoder from the modified reconstructor, followed by three additional convolutional layers, as illustrated in \cref{fig:model_architectures}.c.
In the figure, we display the decoder and its incoming skip connections solely to show how the original encoder–decoder architecture was organized. Model-3 uses only the encoder, and neither the decoder nor its skip connections are part of the model.

\subsubsection{Anchor-free detection head}

We modified YOLOv8 anchor-free decouple head \cite{yolov8_ref} as a couple head and pluged it to the output of backbone networks.
\begin{equation}
    \label{eq:out_channels}
    C_{out} = N_{cls} + 4 \times RegMax
\end{equation}

\begin{figure}[!t]
\centering
\includegraphics[width=0.5\textwidth]{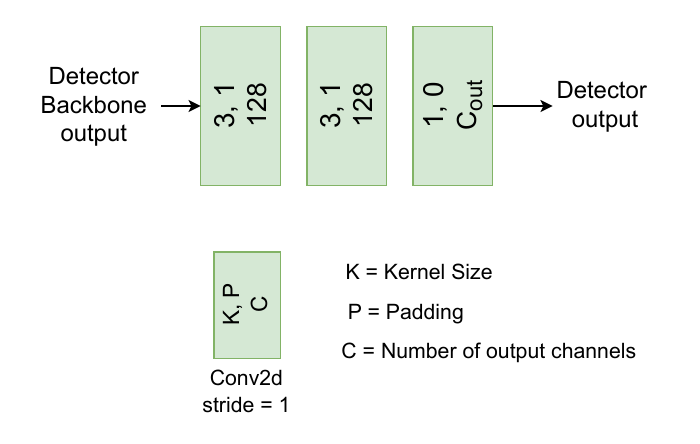}%
\caption{Anchor-free detection head.}
\label{fig:anchorfree_head}
\end{figure}

The layers which belong to Anchor-free detection head is shown in \cref{fig:anchorfree_head}.
The output layer channel count ($C_{out}$) is defined by the number of object classes to be classified ($N_{cls}$) and the discretization levels for a bounding box coordinate regression, termed the regression maximum ($RegMax$) in the Distribution Focal Loss (DFL) \cite{li2021dfl} formulation, as expressed in \cref{eq:out_channels}.

\subsection{ANNs models}

For the ANN models, we used the same architectures as their SNN counterparts, replacing LIF neurons with ReLU activations, as the non-linear activation units. In the case of Model-3, in addition to the stateless fully convolutional version, we also implemented a Conv-LSTM based architecture with added batch normalization layers to facilitate detection performance comparison .

\subsection{Event input encoding}

In our work, we employ two event-encoding methods, a 2D two-channel event histogram \cite{hats} and a voxel-grid \cite{voxelgrid} representation to convert raw event streams from the event-based datasets into tensor formats suitable for model evaluation.

\subsubsection{2D two-Channel event histogram}

The simplest encoding strategy aggregates events within a fixed temporal window into a 3D tensor. The spatial dimensions correspond to the sensor resolution, while the depth dimension consists of two channels, one for positive-polarity events and one for negative-polarity events \cite{hats}. The event histogram is then formed by counting the number of events of each polarity occurring at each pixel location within the temporal window as shown in \cref{fig:encoding_methods}.c.

\subsubsection{Voxel-grid Encoding}

\begin{figure*}[!t]
\centering
\includegraphics[width=1.1\textwidth]{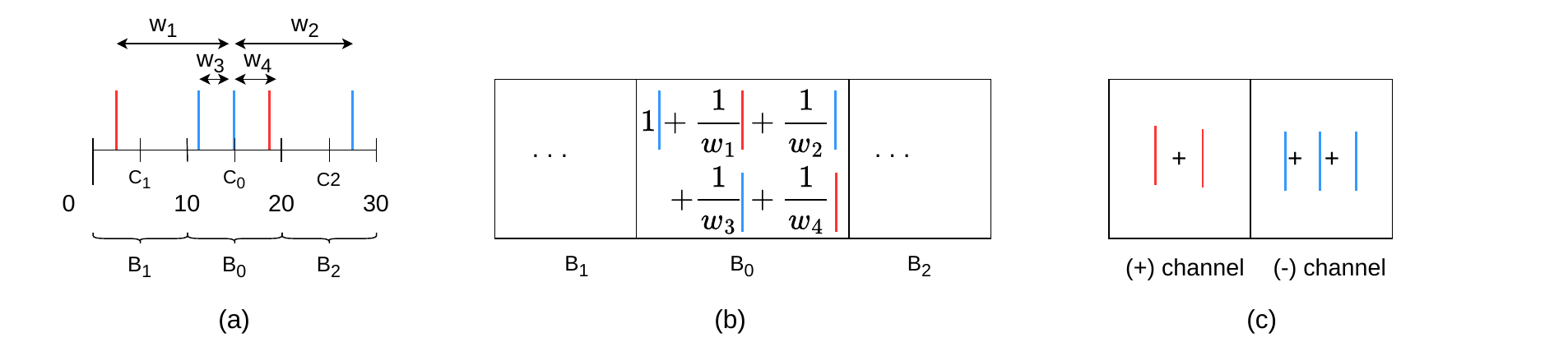}%
\caption{\textcolor{black}{Event-based data encoding for event-based object detection: (a) Positive (red) and negative (blue) events triggered within a 30 ms time period. (b) A Voxel-grid representation with three bins ($B_{0}$, $B_{1}$, and $B_{2}$) each with a temporal length of 10ms. It illustrates how the value of bin $B_{0}$ is computed using weights based on temporal distances from its center ($C_{0}$). (c) A 2D two-channel event histogram, where all positive events are accumulated into one bin and negative events into another. Positive events are generated when the perceived intensity at a pixel increases, while negative events are generated when the perceived intensity decreases.}}
\label{fig:encoding_methods}
\end{figure*}

Alternatively, a richer temporal representation can be obtained using a voxel-grid encoding. Given $N$ input events and $B$ temporal bins, event timestamps are first normalized to the range $[0,B - 1]$.
Events are then distributed across the temporal bins using linear interpolation, producing a 3D tensor with $B$ channels that captures both spatial and temporal event structure.
This approach, described in \cite{voxelgrid}, preserves temporal dynamics more effectively than simple histogram method. A simple graphical illustration for Voxel-grid encoding is shown in \cref{fig:encoding_methods}.b for  events occued in a 30ms period \cref{fig:encoding_methods}.a.

\subsection{ANN-to-SNN Knowledge Distillation}

\begin{figure*}[!t]
\centering
\includegraphics[width=1.15\textwidth]{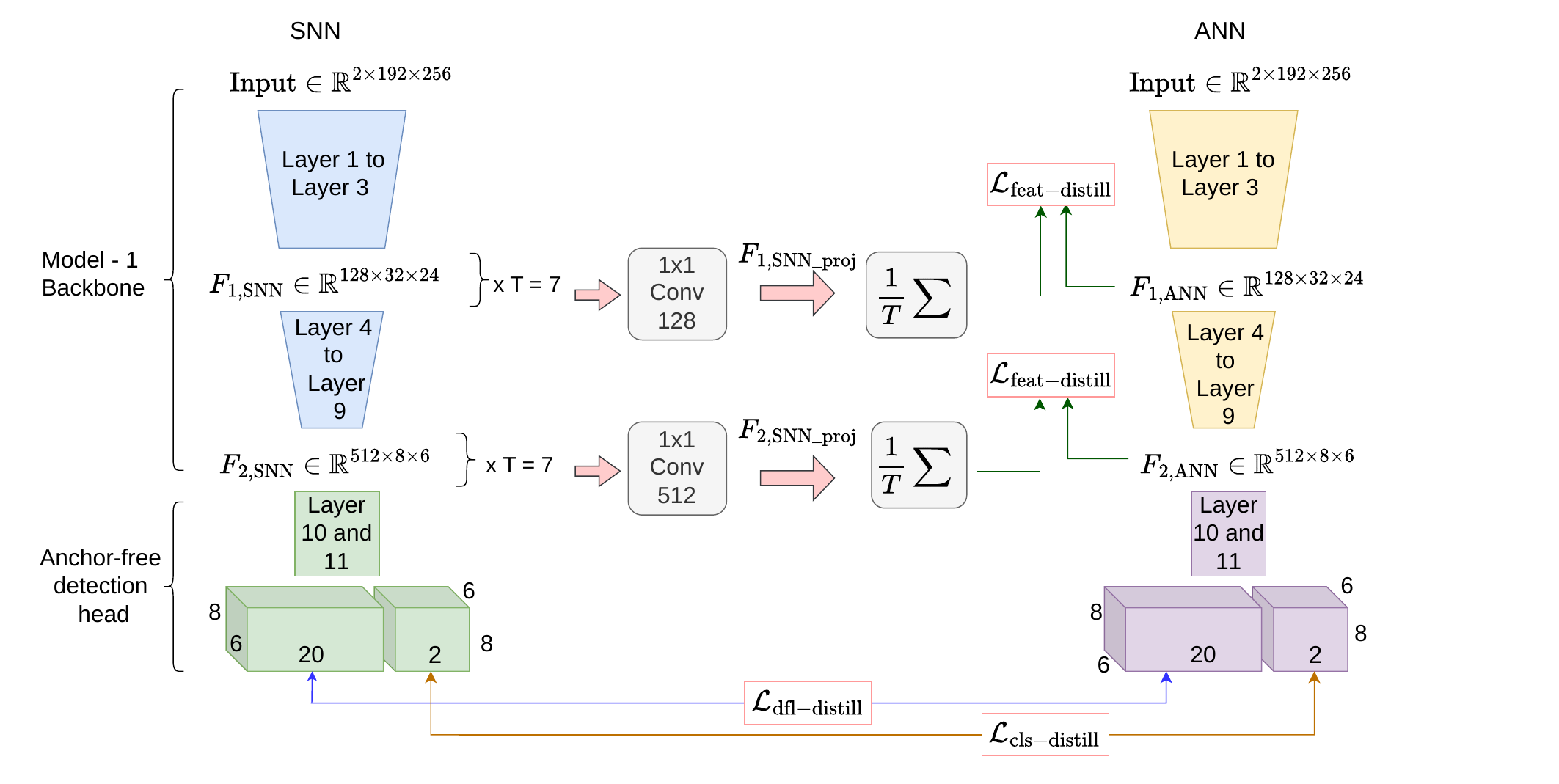}%
\caption{\textcolor{black}{ANN to SNN distillation for Model-1 for prophesee GEN1 dataset which contains two different object classes: SNN Feature maps from 3rd layer and 9th layer are convolved with learnable 1x1 convolutions and then take the average across time dimesion 'T' ( T = 7 in our case). Then the feature map distillation loss ($\mathcal{L}_{\mathrm{feat-distill}}$) is calculated together with corresponding ANN feature maps. 
Since GEN1 has two object classes to classify 2 channels of output layer feature maps are used to calculate classification loss ($\mathcal{L}_{\mathrm{cls-distill}}$). When $RegMax$ value is 5, giving $ 20\ (4 \times 5)$ feature maps of the output layer are used to calculate regression distillation loss ($\mathcal{L}_{\mathrm{dfl-distill}}$).}}

\label{fig:distillation_diagram}
\end{figure*}

We perform ANN-to-SNN distillation using three complementary components: intermediate feature map distillation, classification output logit distillation, and regression output logit distillation for bounding box localization.

\textcolor{black}{First, feature-level distillation transfers knowledge from the ANN’s intermediate feature maps to the corresponding SNN representations. To align SNN feature maps $F_{SNN}$ with the teacher ANN maps $F_{ANN}$, each $F_{SNN}$ is first projected into a suitable space using a learnable $1 \times 1$ convolution, producing $F_{SNN\_proj}$. This projection compensates for mismatches between SNN and ANN feature spaces that arise from differences in their numerical representations, as the learnable convolution can adapt its parameters to adjust the SNN feature maps to better align with the ANN feature maps. Finally, the projected SNN feature map is averaged over the temporal dimension as \[ \bar{F}_{SNN\_proj} = \frac{1}{T} \sum_{t=1}^{T} F_{SNN\_proj}[t].\]}
Then the intermediate feature map distillation loss is defined as in \cref{eq:feat_distillation}.

\begin{equation}
\mathcal{L}_{\mathrm{feat-distill}} = \frac{1}{B H W C} \, \lVert F_{ANN} - \bar{F}_{SNN\_proj} \rVert_F^2
\label{eq:feat_distillation}
\end{equation}

Second, classification-level distillation aligns the ANN and SNN output logit feature maps associated with class prediction. When the Batch size is $B$ and $T_{p}$ is predefined constant, the classification distillation loss can be defined as in \cref{eq:classi_distill} where $N_{cls}$ (defined in \cref{eq:out_channels}) denotes the number of classification output logit feature maps, also equal to the number of target classes.
In both \label{eq:classi_distill} and \label{eq:reg_distill}, $D_{KL}$ refers to the KL divergence loss, which is defined in \cref{eq:L_KL} of \cref{appendix:d}.

\begin{equation}
    \mathcal{L}_{\mathrm{cls-distill}} = \frac{{Tp}^2}{B} \sum{D_{\mathrm{KL(P_{SNN} || P_{ANN})}_{K = N_{cls}}}}
    \label{eq:classi_distill}
\end{equation}

Third, bounding box coordinates regression distillation loss aligns the ANN and SNN output feature maps used for bounding box regression. When the batch size is $B$ and $N$ is the number of bounding boxes per sample, the regression distillation loss can be defined as in \ref{eq:reg_distill} where the number of regression output logit feature maps is $4 \times RegMax$ also defined in \cref{eq:out_channels}.

These distillation losses are graphically illustrated in \cref{fig:distillation_diagram} considering ANN to SNN distillation for Model-1 for prophesee GEN1 dataset which has two object classes.

\begin{equation}
\label{eq:reg_distill}
\mathcal{L}_{\mathrm{dfl-distill}} = \frac{{Tp}^2}{B N} 
\sum_{b=1}^{B} \sum_{n=1}^{N} 
D_{\mathrm{KL(P_{SNN} || P_{ANN})}_{K = 4 \times RegMax}}
\end{equation}

The total loss $\mathcal{L}_{tot}$ which is defined in \cref{eq:distill_detect_loss} balances the YOLOv8 detection losses ($L_{box}$, $L_{dfl}$ and $L_{cls\text{-}distill}$ which are defined in \cref{eq:yolov8_box}, \cref{eq:yolov8_dfl} and \cref{eq:yolov8_cls} of \cref{appendix:d} respectively.) and distillation loss components via hyperparameters $\alpha, \beta, \gamma, \theta,$ and $\eta$.
The coefficients $\alpha, \beta, \gamma, \theta,$ and $\eta$ control the relative contribution of each loss component to the overall loss.

\begin{equation}
\label{eq:distill_detect_loss}
\begin{aligned}
L_{tot} &= \alpha L_{box} + \beta L_{cls} + \gamma L_{dfl} \\
        &\quad + \theta L_{cls\text{-}distill}
        + \theta L_{dfl\text{-}distill}
        + \eta L_{feat\text{-}distill}
\end{aligned}
\end{equation}

\subsection{Evaluation Metrics}

To evaluate detection performance, we adopt standard COCO metrics \cite{coco}, including mAP$_{0.5}$ to assess detection accuracy at an IoU threshold of 0.5,
mAP$_{0.5:0.95}$ to capture localization performance under increasing stringency, and F1$_{iou@0.5}$ to balance precision and recall.

For neuromorphic hardware evaluation on Loihi 2, we follow established profiling methodologies \cite{loihi_profile_tute, SDNN_Loihi2} define the inference rate (or throughput) ($R$), per inference energy ($E$), latency ($L$), and Energy-Delay Product ($EDP$) for SNNs as:
\begin{equation}
\label{eq:hardware_metrics}
    R = \frac{1}{\Delta t \cdot T}, \quad E = \frac{P_{avg}}{R}, \quad L = \Delta t \cdot N_{layer}, \quad EDP = L \times E
\end{equation}
where $\Delta t$ is the average time per simulation time step, $T$ is the total time steps, $P_{avg}$ is the power (either static, dynamic or total) from the Loihi 2 profiler, and $N_{layer}$ is the number of deployed layers. For ANNs, latency is simply $1/R$.
For ANN baselines, latency is simply $1/R$ which is the inverse of the inference rate.

\paragraph*{Sparsity and Synaptic Operations (SOPs)} For a SNN model with $N$ neurons and $T$ timesteps over $M$ samples, let $S_i$ be the spikes emitted by neuron $i$. Sparsity ($S$) and the resulting SOPs per inference are defined as:
\begin{equation}
\label{eq:sop_metrics}
    S = \frac{1}{NM} \sum_{i=1}^{N} S_i, \quad SOPs = S \times N_{\text{spk\_neurons}}
\end{equation}
where $N_{\text{spk\_neurons}}$ is the count of LIF neurons with a spiking threshold.




\section{\textbf{Setup}}

In this section, we describe our experimental setup, including the tools, parameters, and frameworks used for implementing and training both ANN and SNN models. We also detail the datasets and preprocessing strategies, as well as the complete setup used to benchmark the performance of the SNNs and ANNs.

\subsection{Datasets and data preprocessing}

We primarily used two event-based datasets and two frame-based datasets to benchmark our models. In addition, we introduce a new frame-based video dataset, referred to as the UAV-based Tunnel Inspection Dataset.

\subsubsection{Event-based datasets}

The first event-based dataset is the event-based component of ev-CIVIL \cite{Gamage2025_evCIVIL}, which focuses on detecting two types of civil-infrastructure defects which are cracks and spalling,both of which exhibit irregular and non-uniform shapes. These defects are treated as anomaly classes. Throughout the paper, we refer to this event-based component as evCIVIL-ev. The dataset provides a single training set and two distinct test sets: one captured under adequate illumination and another under low-light conditions.
For evaluating detection accuracy, each test set was used independently. However, for benchmarking power consumption, energy, latency, and inference rate, samples from both test sets were combined, as separate benchmarking yielded negligible differences.

The second event-based dataset is Prophesee GEN1 \cite{gen1}, which contains two object classes: cars and pedestrians.
Unlike evCIVIL-ev, these objects have regular and well-defined shapes, making GEN1 a complementary benchmark for evaluating performance on structured, conventional detection tasks.

\subsubsection{Frame-based datasets}

The first frame-based dataset is the frame-based component of ev-CIVIL \cite{Gamage2025_evCIVIL}, captured simultaneously with the event stream. We refer to this component as evCIVIL-fr.

As the second frame-based dataset, we used PASCAL VOC \cite{pascal_voc}, a widely used benchmark consisting of 20 object classes.

\subsubsection{UAV-based Tunnel Inspection Dataset}

We introduce a new grayscale dataset collected via UAV during the inspection of an abandoned tunnel in Huesca, Spain,focusing on the tunnel’s interior ceiling and side walls.
A DJI Matrice 100 quadrotor equipped with two cameras was used for data acquisition: a 4 MP IDS UI-3251LE-M-G camera with a 4 mm lens, referred to as the up camera, was mounted facing upward to record the ceiling, while a 12 MP Basler acA4112-20um camera with an 8 mm lens, referred to as the right camera, was positioned to capture the side walls at an approximate distance of 3 meters. Both cameras operated at 15 frames per second. To ensure consistently illuminated imagery regardless of the tunnel’s ambient lighting, the UAV was fitted with four shutter-synchronized LEDs aligned with each camera’s field of view, operating at 15 Hz with an energy-efficient duty cycle of about 5\%. Two manual flights were conducted along the selected tunnel , with lengths of 58.53 meters for flight-1 and approximately 134.5 meters for flight-2, while the tunnel itself measured around 7.4 meters in width and 4.75 meters in height.

\begin{figure}[htbp]
\centering
\includegraphics[width=0.8\textwidth, trim=0 10 0 10, clip]{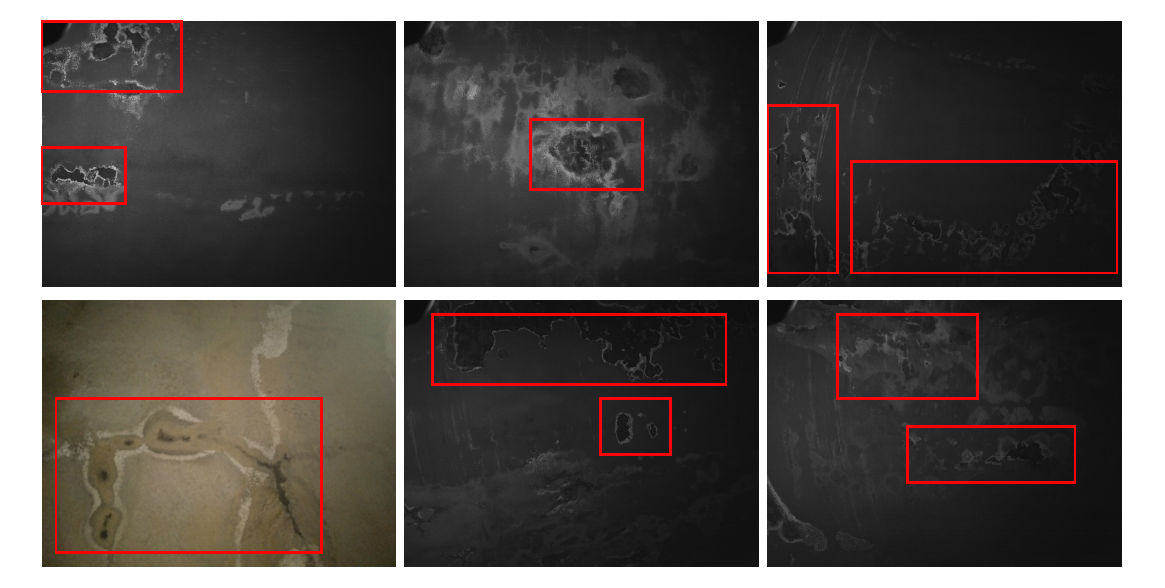}

\vspace{-3mm} 
\caption{\textcolor{black}{Sample images from the tunnel inspection dataset with bounding box–annotated defects}}
\label{fig:tunnel_example_imgs}
\vspace{-4mm} 
\end{figure}

A total of 2,779 images from the up camera (1600×1200 resolution) and 3,308 images from the right camera (4112×2780 resolution) were collected. To reduce redundancy, frames were down-sampled to 5 frames per second. Tunnel defects—including concrete loss, missing wall patches, and water-leakage damage—were manually annotated as a single class labeled “defects,” using the standard YOLO \cite{YOLOv3_ref} format. In total, 4,234 bounding boxes were annotated in the up-camera images and 5,207 in the right-camera images.
Some example images with bounding box annotations for the available defects are shown in \cref{fig:tunnel_example_imgs}.

For evaluation, 660 right-camera images and 555 up-camera images were randomly selected for the test set, while 330 right-camera and 270 up-camera images formed the validation set, with the remaining images used for training. The dataset is publicly available at the link provided earlier.

\subsubsection{Data preprocessing}

To evaluate detection performance,, We used the same train, validation and test sets used to evaluate ev-CIVIL \cite{Gamage2025_evCIVIL},prophesee GEN1 \cite{gen1} and PASCAL VOC \cite{pascal_voc} datasets.
For GEN1, each sample consists of events within a period of 100ms and for evCIVIL-ev each sample consist of events within a time window whose window length is dynically determined as in \cite{Gamage2025_evCIVIL}.
The events of those samples were encoded as tensors using the simplest 2D two-channel histogram representation to evaluate the performance of Model-1 and Model-2.

However, for Model-3, we employed a three-channel voxel-grid representation, as we used the same voxel-grid–based dataset as \cite{rebecq} to pretrain the model and initialize its weights prior to fine-tuning. For fairness, we also evaluated Model-1 and Model-2 with the voxel-grid encoding; however, this configuration did not surpass their performance obtained using the 2D two-channel event-histogram representation. In contrast, Model-3 achieved higher detection accuracy with the voxel-grid representation than with the histogram encoding, which we attribute to its pretrained initialization based on the dataset used in \cite{rebecq}. Therefore, throughout the paper, we report results for Model-1 and Model-2 using the 2D two-channel histogram representation, and for Model-3 using the voxel-grid representation.

For frame-based images preprocessing, pixel values were normalized to the [0,1] range dividing by 255. Images in evCIVIL-fr were fed as single-channel grayscale inputs, while PASCAL VOC images were provided as three-channel color images.

The input resolution for GEN1, evCIVIL-ev, and evCIVIL-fr datasets was set to 256 × 192 pixels to preserve the native aspect ratio and minimize spatial distortion while maintaining comparable computational complexity.
For both PASCAL VOC and the UAV-based Tunnel Inspection Dataset, all images were resized to 224 × 224 pixels to match the standard input dimensions commonly used in CNN architectures.
\Cref{tab:model_complexity} shows the number of activation neurons and the number of parameters for the three models based on the specified input resolutions. It can be observed that the number of neurons in all models remains below 1 million, indicating that all models are deployable on a single Loihi 2 chip.

\begin{table}[htbp]
\centering
\caption{Neuron and Parameter Counts for Each Model at Targeted Input Resolutions.}
\label{tab:model_complexity}
\footnotesize
\setlength{\tabcolsep}{0pt}
\begin{tabular*}{\columnwidth}{@{\extracolsep{\fill}} l cc cc}
\hline
\multirow{2}{*}{Model} & \multicolumn{2}{c}{Resolution: 256x192} & \multicolumn{2}{c}{Resolution: 224x224} \\
\cline{2-3} \cline{4-5}
 & \# Neurons & \# Params & \# Neurons & \# Params \\
\hline
Model-1 & 504,864 & 4,775,256 & 516,264 & 4,214,704 \\
Model-2 & 768,748 & 3,845,000 & 787,128 & 3,830,440 \\
Model-3 & 910,752 & 5,619,808 & 929,726 & 5,619,808 \\
\hline
\end{tabular*}
\end{table}

\subsection{Model Implementation and Training}

The SNN models were developed and trained using PyTorch \cite{pytorch} with the SpikingJelly \cite{fang2020spikingjelly} library. The SpikingJelly library provides various spiking neuron models, including the Leaky Integrate and Fire(LIF) neurons that closely mimic the dynamics of the LIF neurons implemented on Intel’s Loihi 2 neuromorphic hardware.

\begin{table}[htbp]
\centering
\caption{SNN Hyperparameters and Training Settings}
\label{tab:snn_hyperparams}
\footnotesize 
\begin{tabular}{ll}
\hline
Parameter & Value \\ \hline
Inference time steps ($T$) & 7 \\
$V_{th}$ of output layer LIFs & 2048 \\
$V_{th}$ of hidden layers LIFs & 1.0 \\
$RegMax$ of anchor-free head & 5 \\ \hline
Batch size & 8 \\
Learning rate (1st stage / Fine-tuning) & 0.001 / 0.0002 \\
Iterations (PASCAL VOC / Other datasets) & 800 / 400 \\
Finetuning iterations (mean-only) & 150 \\
Optimizer / Scheduler & AdamW / CosineAnnealLR \\
Batchnorm mean clamp range & -0.98 to 1.1 \\ \hline
\multicolumn{2}{l}{$V_{th}$: neuron threshold voltage}
\end{tabular}
\end{table}

\textcolor{black}{In our SNNs, the threshold voltage of the hidden-layer LIF neurons was set to 1, while the output-layer LIF neurons were assigned a much higher threshold of 2048. This prevents the output neurons from spiking, allowing their membrane voltages to accumulate and be used directly for bounding box regression and classification logits. The $RegMax$ parameter (defined in \cref{eq:out_channels}) was empirically set to 5, balancing detection accuracy with the number of output channels—higher values can improve accuracy but require more output ports. \Cref{tab:snn_hyperparams} summarizes the key hyperparameters, which were empirically chosen and applied to our SNN models and training procedures.}

\subsubsection{Model Initialization and Training Procedure}

The overall workflow for model development, training, deployment, and benchmarking on Loihi 2 is illustrated in \cref{fig:overall_setup}.
We chose the SNN direct training approach over ANN-SNN conversion due to its advantage of reduced inference latency. 
SNN training was performed using the Back Progation Through Time (BPTT) \cite{snn_backprop} algorithm implemented in SpikingJelly \cite{fang2020spikingjelly}.

For frame-based datasets, all ANNs were initialized with random weights. For event-based datasets with paired image data (e.g., ev-CIVIL), event-driven models were initialized using pretrained weights from their corresponding frame-based ANNs to accelerate convergence. In cases without paired data, event-based ANNs were initialized using weights acquired from general frame-based pretraining.
For directly trained SNNs (Model-1 and Model-2), Kaiming normal initialization was standard, though pretrained ANN weights were used for layer-wise initialization when available. For Model-3, the first seven layers utilized pretrained weights from an event-to-image reconstructor \cite{rebecq}, while the remaining layers followed a Kaiming normal distribution.

For object detection, we employ standard YOLOv8 loss functions. The event-to-image reconstructor in Model-3 is trained using Learned Perceptual Image Patch Similarity (LPIPS) and temporal consistency losses \cite{rebecq}, with the latter calculated using optical flow estimated by the RAFT model \cite{teed2020raft}. All SNNs undergo 8-bit Quantization-Aware Training (QAT) via the SpikingJelly library to ensure high-fidelity deployment on Loihi 2.

When leveraging knowledge distillation for Models 1 and 2, we incorporate a teacher-student framework. Following the procedure for direct SNN training, the objective function in \cref{eq:distill_detect_loss} augments the standard YOLOv8 loss with weighted distillation components, aligning the spiking student’s feature maps and predictions with those of the ANN teacher.

In our work, for all the distillation loss functions, we set the temperature ($T_p$ to 20, following the setting used in \cite{li2022yolov6}.
In \cref{eq:distill_detect_loss}, we set $\beta = 0.5$ and $\gamma = 1.5$ following \cite{yolov8_ref}. The distillation weight $\theta$ was scheduled using a cosine decay from 1 to 0 over the training iterations, as in \cite{li2022yolov6}. For the feature-distillation weight $\eta$, we used a value of 1 for the first 20\% of iterations and 0.01 for the remaining iterations, based on empirical observations. The coefficient $\alpha$ was set to 6.0, also determined empirically.
Knowledge distillation was not applied to Model-3, since its fully convolutional ANN counterpart did not exhibit a significant performance advantage compared to the directly trained SNN version.

All models were trained on a workstation equipped with an NVIDIA RTX 4090 GPU with 24\ GB VRAM.

\subsubsection{Inference Time Steps and Reset Strategy}

Since the trained SNNs are ultimately deployed on Loihi 2, their inference configuration must respect several hardware-specific constraints. In particular, Loihi 2 requires periodic layer resets to maintain stable neural states.
These resets must occur at intervals of the form $2^{x}$ where $x$ is a positive integer ($x \geq 1$).
Accordingly, the reset interval was set to 8, which was found empirically to offer a suitable trade-off between accuracy and runtime efficiency.

During inference, the network output is decoded every 7 time steps, with the $8^{th}$ time step reserved for neuron reset and internal state maintenance (i.e., housekeeping).
When the model is executed on Loihi 2, results are read out with an initial offset of $N+2$ where $N$ is the number of layers in the deployed SNN.
The additional offset of 2 accounts for initial data-loading and buffering overhead.

\subsubsection{Reparameterization and deployment on Loihi 2}

As shown in \cref{fig:overall_setup}, after quantization-aware training of the SNN models, the RepVGG \cite{repvgg} reparameterization technique was used to convert residual blocks into equivalent feedforward blocks. Then, the BatchNorm variances were absorbed into the preceding convolution weights and BatchNorm means, resulting in a mean-only BatchNorm form, as Loihi 2 currently supports only mean-only batch normalization \cite{lava_dl_pilotnet_sdnn}.

\textcolor{black}{Next, the models were fine-tuned for 150 iterations with an initial learning rate of 0.0002, using mean-only batch normalization. Since the batch normalization mean is incorporated as a negative bias in leaky integrate-and-fire (LIF) neurons, excessively low mean values can lead to unstable behavior. In particular, if the BatchNorm mean decreases to $\leq -1$, neurons may spike continuously because the LIF neuron threshold voltage is 1.
To prevent this, during fine-tuning we apply a lower bound constraint on the mean value by clamping it to $-0.98$. This ensures that the BatchNorm mean remains $\geq -0.98$, thereby avoiding continuous spiking and maintaining stable network dynamics.}

Finally, the SNNs developed on SpikingJelly were exported to Lava-DL–compatible models and then converted into Loihi 2's '.net' format so that they can be run on Loihi 2.

\subsection{Benchmarking process}

We benchmarked the detection performance of our models on Loihi 2 and compared the results against two edge GPUs (Jetson nano) and an energy-efficient CPU (Apple M2).

To ensure statistical reliability and alignment with established neuromorphic benchmarking protocols we randomly selected 20 samples from each test dataset and executed 5 benchmarking iterations per sample \cite{loihi_profile_tute},\cite{SDNN_Loihi2}. All reported metrics including power, energy, latency, inference rate, and Energy-Delay Product (EDP) represent the resulting average values across all samples and iterations. Consistent with existing Loihi 2 literature, we intentionally omit standard deviation reporting to maintain alignment with state-of-the-art results. By applying this 20-sample, 5-iteration procedure across all three platforms, we established a consistent baseline for our comparisons.

We used Intel Oheo Gulch \cite{loihi2_oheo_gulch}, Loihi 2 single-chip development board based on Intel’s Loihi 2 processor, to benchmark spiking neural networks (SNNs). Access to Oheo Gulch was provided remotely.
To run on Loihi 2, all data samples were converted from PyTorch tensors to NumPy arrays for evaluation on the SNN models running on Loihi 2. Each data sample was formatted with dimensions W × H × C, where the first two correspond to the spatial dimensions (width and height) and the last corresponds to the channel dimension.
In profiling the SNNs on Loihi 2, We employed the Loihi 2 profiler as described in \cite{loihi_profile_tute} to extract power consumption, latency, inference rate, and energy usage, while detection accuracy was evaluated using the Lava-dl model outputs.
Following Intel’s guidance, all benchmarking results were obtained excluding I/O (Input/Output) overhead, since high-speed I/O support for Loihi 2 is still under development.

We used two NVIDIA Jetson Nano devices in our experiments. First, the Jetson Nano B01 module, integrated with a Yahboom Jetson Nano Development Board \cite{yahboom_jetson_nano},, was employed to benchmark artificial neural networks (ANNs) on an edge GPU platform, leveraging NVIDIA TensorRT for optimized inference.

Subsequently, we utilized the more advanced NVIDIA Jetson Orin Nano 8\ GB development kit, which offers improved energy efficiency and significantly higher throughput, making it better suited for demanding edge AI workloads \cite{nvidia_jetson_orin_nano_super}.

We used two Jetson nano devices.
First, the Jetson Nano B01 module, integrated with a Yahboom development board \cite{yahboom_jetson_nano}, was used to benchmark ANNs on an edge GPU platform, leveraging NVIDIA TensorRT for optimized inference.
Then the cutting Nvidia Jetson Orin Nano 8\ GB with Development board more optimzed for energy efficiency and high throughput. 
On both of these devices,the benchmarking procedure followed the methodology in \cite{SDNN_Loihi2}, which uses tegrastats to obtain the Jetson’s memory usage and power consumption in real time.

\textcolor{black}{Before deployment on Jetson Nano, the PyTorch-based ANN models were exported to the ONNX intermediate representation and underwent platform-specific optimization using NVIDIA TensorRT to run on it's edge GPU.
We utilized FP16 precision to leverage the hardware's specialized floating-point capabilities, significantly reducing inference latency compared to standard FP32.
To ensure a fair and consistent comparison between platforms, we clarify that neither Loihi 2 nor Jetson Nano evaluations were conducted in a fully end-to-end manner with external I/O pipelines.
As in Loihi 2 on the Oheo Gulch development board, for the Jetson Nano also,input samples were preloaded into memory, and inference was performed directly from these in-memory data without involving external data transfer during runtime.
So here first we load same randomly selected input data samples (as for Loihi 2) into NumPy array, and inference was executed directly from memory without involving external data transfer or peripheral I/O during runtime.
To isolate compute-related energy, we first measured the idle power of each system with all major components active (CPU, GPU, memory, and I/O subsystems) but without model execution. We then measured the total power and energy during inference over multiple runs (2048 runs over randomly selected samples).
The inference rate (throughput) was measured based on the total time taken to process multiple runs (2048 runs over randomly selected samples) using these preloaded inputs.
The dynamic energy was computed as the difference between total energy and idle energy, ensuring that only the energy associated with model execution is considered. This methodology consistently accounts for baseline system components, including memory access and internal data movement, on both platforms.}

For CPU performance comparison, we used a MacBook Air equipped with an ARM-based Apple Silicon M2 chip \cite{MacRumors_M2_2022}, one of the most energy-efficient laptops available. It offers strong computational performance while consuming significantly less power than comparable x86-based systems, particularly for deep learning inference. Since the system relies solely on the CPU, power measurements primarily reflect CPU usage. We used the built-in powermetrics tool on the MacBook’s to measure power consumption, inference latency, and overall energy usage.
In this study, energy values calculated from the net main loop power measurements provided by the powermetrics tool were used as both dynamic and total energy.

\begin{table*}[t!]
\centering
\caption{Detection performance of directly trained SNN models vs. ANN baselines. For each test set, the highest value for each metric across all models (both ANN and SNN implementations) is highlighted.( '--' denotes to not applicable)}
\label{tab:accuracy_direct_train}
\fontsize{8pt}{8pt}\selectfont
\setlength{\tabcolsep}{2.5pt}
\begin{tabular}{l|ccc|ccc|ccc|ccc|ccc|ccc}
\toprule
 & \multicolumn{6}{c|}{evCIVIL-fr} & \multicolumn{6}{c|}{evCIVIL-ev} & \multicolumn{3}{c|}{\multirow{2}{*}{PASCAL VOC}} & \multicolumn{3}{c}{\multirow{2}{*}{GEN1}} \\
 & \multicolumn{3}{c}{Day} & \multicolumn{3}{c|}{Night} & \multicolumn{3}{c}{Day} & \multicolumn{3}{c|}{Night} & \multicolumn{3}{c|}{} & \multicolumn{3}{c}{} \\
Model & M1 & M2 & F1 & M1 & M2 & F1 & M1 & M2 & F1 & M1 & M2 & F1 & M1 & M2 & F1 & M1 & M2 & F1 \\
\midrule
\textit{Model-1} & & & & & & & & & & & & & & & & & & \\
ANN & .43 & .23 & .38 & .08 & .07 & .09 & .43 & .21 & .41 & .44 & \textbf{.19} & .43 & .45 & .24 & .36 & .45 & .24 & .36 \\
SNN & .39 & .20 & .37 & .08 & .04 & .07 & .36 & .16 & .37 & .37 & .15 & .33 & .40 & .23 & .36 & .40 & .19 & .36 \\
\hline
\textit{Model-2} & & & & & & & & & & & & & & & & & & \\
ANN & \textbf{.45} & \textbf{.23} & \textbf{.41} & .11 & \textbf{.07} & \textbf{.07} & \textbf{.44} & \textbf{.22} & \textbf{.43} & \textbf{.45} & .18 & \textbf{.47} & \textbf{.49} & \textbf{.31} & \textbf{.42} & .45 & .23 & .37 \\
SNN & .40 & .20 & .37 & .09 & .07 & .04 & .38 & .16 & .38 & .39 & .15 & .35 & .42 & .25 & .35 & .42 & .22 & .37 \\
\hline
\textit{Model-3} & & & & & & & & & & & & & & & & & & \\
ANN & -- & -- & -- & -- & -- & -- & .51 & .28 & .48 & .53 & .23 & .55 & -- & -- & -- & .49 & .27 & .44 \\
SNN & -- & -- & -- & -- & -- & -- & .46 & .21 & .45 & .48 & .19 & .49 & -- & -- & -- & .45 & .25 & .44 \\
\hline
\textit{Model-3-mem} & & & & & & & & & & & & & & & & & & \\
ANN & -- & -- & -- & -- & -- & -- & .39 & .20 & .38 & .40 & .16 & .38 & -- & -- & -- & \textbf{.47} & \textbf{.28} & \textbf{.36} \\
\bottomrule

\multicolumn{8}{l}{\textit{$M1$: $mAP_{0.5}$, $M2$: $mAP_{0.5:0.95}$, $F1$: $F1_{iou@0.5}$}}
\end{tabular}
\end{table*}

\section{\textbf{Experimental Results}}

\textcolor{black}{In this section, we present our experimental results. First, we compare the detection performance of directly trained SNN models with their ANN counterparts. Next, we demonstrate the performance gains achieved through ANN-to-SNN knowledge distillation. We then compare the static and dynamic power consumption of Intel Loihi 2 on the Oheo Gulch development board.}

\textcolor{black}{Subsequently, we evaluate key performance metrics including detection accuracy, inference rate, dynamic energy consumption, and energy-delay product (EDP) across four hardware platforms: Loihi 2, NVIDIA Jetson Nano B01, NVIDIA Jetson Orin Nano, and a MacBook M2 CPU. We also provide a comparative analysis of power consumption across these platforms.}

\textcolor{black}{Furthermore, we benchmark performance on the UAV-based tunnel inspection dataset.
Finally, we analyze the variation of inference rate and dynamic energy with respect to the number of synaptic operations(SOPs) of SNNs during inference.}

\subsection{Detection Performance of Directly Trained Spiking Neural Networks (SNNs)}

In \cref{tab:accuracy_direct_train}, we present the results of the evaluation of models trained directly on frame and event-based datasets. According to the results with respect to frame-based datasets, Model-2 outperforms Model-1 in all three metrics for all test sets and for both the ANN and SNN implementations.
Considering Model-2, the best-performing model, the ANN version consistently outperforms its counterpart SNN by approximately 11–15\% in $mAP_{0.5}$, 15–24\% in $mAP_{0.5:0.95}$ and 10–25\% in $F1_{iou@0.5}$.

For event-based datasets, We evaluate Model-3-mem, an extension of the original Model-3 architecture that incorporates Conv-LSTM memory units as described in \cite{rebecq}. Only the ANN implementation of Model-3-m was tested, providing a baseline indication of any potential performance gap between the memory-augmented model and Model-3.
As per the results in \cref{tab:accuracy_direct_train}, among the four ANN models, Model-3-mem achieves the highest performance on the GEN1 dataset, outperforming the other three models by at least 4.5\% in terms of $mAP_{0.5}$.
In contrast, for the evCIVIL event-based dataset, Model-2 performs the best, exceeding the others by at least 2.5\% in $mAP_{0.5}$.
For SNN-based models, Model-2 is the top performer in event-based datasets. However, it still lags behind its ANN counterpart by 13-27\% in all three metrics.

\begin{figure*}[!t]
\centering
\subfloat[]{\includegraphics[width=0.50\textwidth]{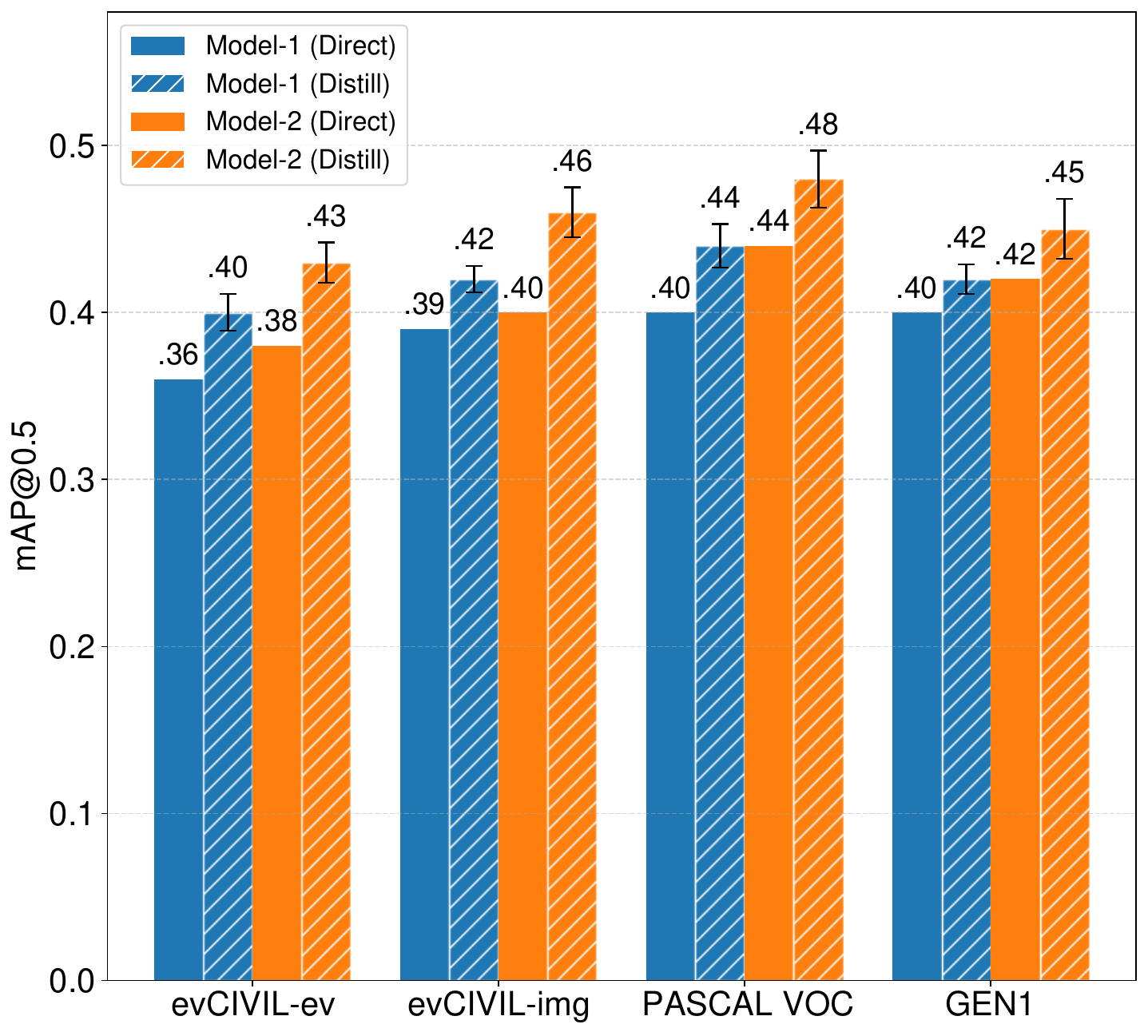}%
\label{fig:distill_first_case}}
\hfil
\subfloat[]{\includegraphics[width=0.49\textwidth]{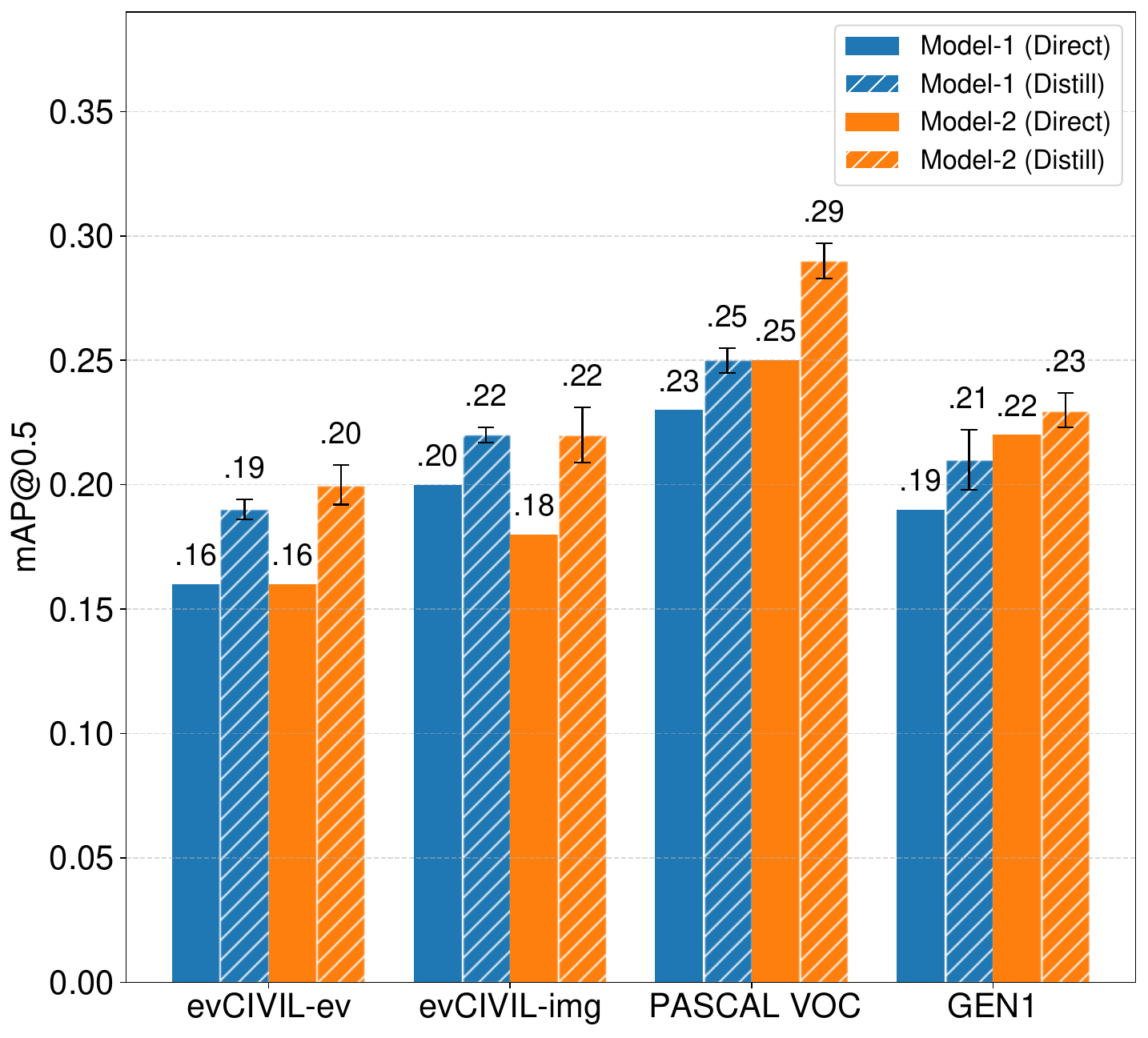}%
\label{fig:distill_second_case}}
\caption{\textcolor{black}{(a) $mAP_{0.5}$ improvements of Model-1 and Model-2 on four benchmark datasets. (b) $mAP_{0.5:0.95}$ improvements of Model-1 and Model-2 on the same datasets. Error bars represent the standard deviation of detection metrics for distilled SNN models. (For each dataset, distillation-aware training was performed five times using different random seeds with Kaiming normal initialization. The bar heights for distilled SNNs indicate the mean performance across these repeated experiments.)}}
\label{fig:distill_performance}
\end{figure*}

\subsection{\textcolor{black}{Effect of ANN-to-SNN Knowledge Distillation on SNNs' Detection Performance}}

The \cref{fig:distill_performance} compares the average detection accuracy of distilled SNN models with directly trained SNNs in terms of $mAP_{0.5}$ (\cref{fig:distill_first_case}) and $mAP_{0.5:0.95}$ (\cref{fig:distill_second_case}) across both Model-1 and Model-2. The figure also presents the standard deviation of detection accuracy as error bars.

To assess training stability, distillation-aware training was repeated five times for each dataset, with each run initialized using different random seeds under Kaiming normal initialization. The results demonstrate high stability, with standard deviations ranging from $0.003$ to $0.012$ for $mAP_{0.5:0.95}$, and from $0.008$ to $0.017$ for $mAP_{0.5}$ across all datasets and models.

Based on the mean accuracy values of the distilled SNN models, several observations can be made. As shown in \cref{fig:distill_first_case}, for frame-based datasets, distillation yields performance gains of $3\text{--}14\%$ in $mAP_{0.5}$, while for event-based datasets, improvements range from $5\text{--}13\%$ across both models. Similarly, \cref{fig:distill_second_case} indicates that, in terms of $mAP_{0.5:0.95}$, distilled models achieve gains of $10\text{--}12\%$ on frame-based datasets and $5\text{--}23\%$ on event-based datasets. Among all variants, Model-2 consistently achieves the best performance.

In \cref{fig:distill_performance}, `evCIVIL-ev' and `evCIVIL-fr' correspond to the evCIVIL\_ev\_day and evCIVIL\_fr\_day subsets, respectively. Although not shown in the figure, the evCIVIL\_ev\_night subset also exhibits improvements of $8\%$ in $mAP_{0.5}$ and $13\%$ in $mAP_{0.5:0.95}$.

Overall, the distilled SNN models retain $87\text{--}100\%$ of the detection performance of their corresponding ANN counterparts in terms of $mAP_{0.5:0.95}$, and $93\text{--}98\%$ in terms of $mAP_{0.5}$ across all evaluated datasets.
Some qualitative visualizations of the detection results obtained by the SNN variant of Model-2, across both frame-based and event-based datasets, are provided in \cref{fig:qualititative_visualize_4datasets} (see \cref{appendix:a}).

\subsection{\textcolor{black}{Comparison of Static vs Dynamic Power consumption of Loihi 2 on Oheo Gulch}}

\begin{table}[htbp]
\centering
\caption{\textcolor{black}{Static and dynamic power consumption (W) for a single Loihi 2 chip on Oheo Gulch development board (n/a refers to 'Not Applicable')}}
\label{tab:power_comparison}
\footnotesize 
\setlength{\tabcolsep}{0pt} 
\begin{tabular*}{\columnwidth}{@{\extracolsep{\fill}} l cc cc cc}
\hline
\multirow{2}{*}{\textbf{Dataset}} & \multicolumn{2}{c}{\textbf{Model-1}} & \multicolumn{2}{c}{\textbf{Model-2}} & \multicolumn{2}{c}{\textbf{Model-3}} \\
\cline{2-3} \cline{4-5} \cline{6-7}
 & static & dynamic & static & dynamic & static & dynamic \\
\hline
evCIVIL-ev & 1.71 & 0.38 & 1.72 & 0.67 & 1.74 & 0.50 \\
evCIVIL-fr & 1.81 & 0.35 & 1.74 & 0.82 & n/a & n/a \\
PASCAL VOC & 1.71 & 0.34 & 1.74 & 0.71 & n/a & n/a \\
GEN1 & 1.83 & 0.37 & 1.70 & 0.77 & 1.74 & 0.64 \\
\hline
\end{tabular*}
\end{table}

\Cref{tab:power_comparison} presents the static and dynamic power consumption of Loihi~2. As Loihi~2 is integrated within the Oheo Gulch development board, the reported static power corresponds to the baseline power consumption of the board when no SNN models are running, which includes the leakage power of Loihi~2 also. The dynamic power represents the additional power consumption incurred during the execution of SNN models on Loihi 2 chip.
Across all models and datasets, static power consumption on Loihi 2 is significantly higher than dynamic power. For example, in Model-2, static power exceeds dynamic power by more than 2$\times$, whereas in Model-1, static power is approximately 4.5–5.2$\times$ higher than dynamic power. Similarly, for event-based datasets, the static power of Model-3 is observed to be 2.5–3.5$\times$ higher than its dynamic power. A comprehensive comparison of static and dynamic power across all four datasets and three models is provided in the \cref{tab:power_comparison}.

\subsection{Comparison of Inference Rate and Dynamic Energy consumption: Loihi 2 vs. NVIDIA Jetson nano and MacBook M2 CPU}

\begin{figure*}[!t]
\centering
\subfloat[]{\includegraphics[width=0.50\textwidth]{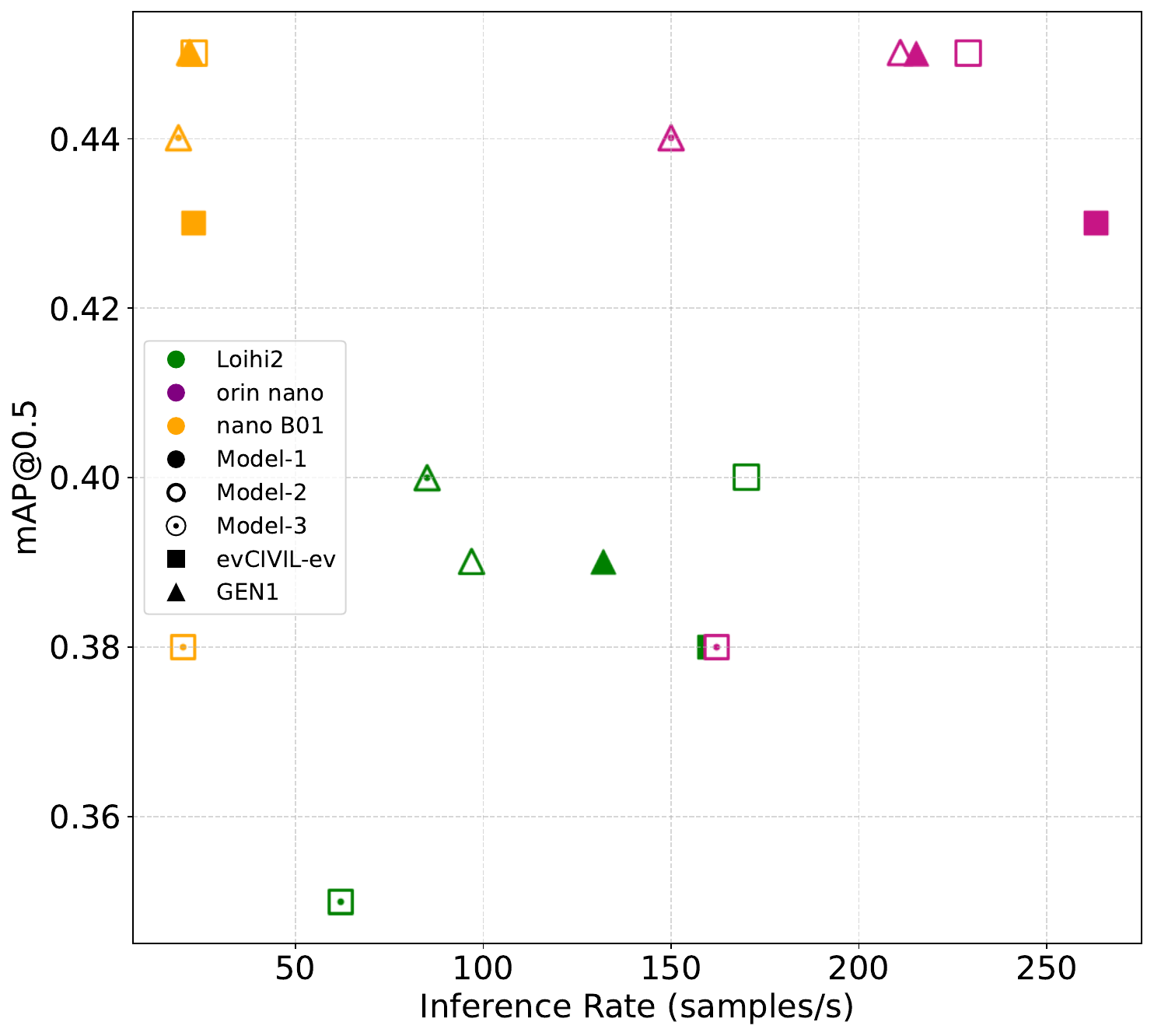}%
\label{fig:inf_rate_event_based}}
\hfil
\subfloat[]{\includegraphics[width=0.49\textwidth]{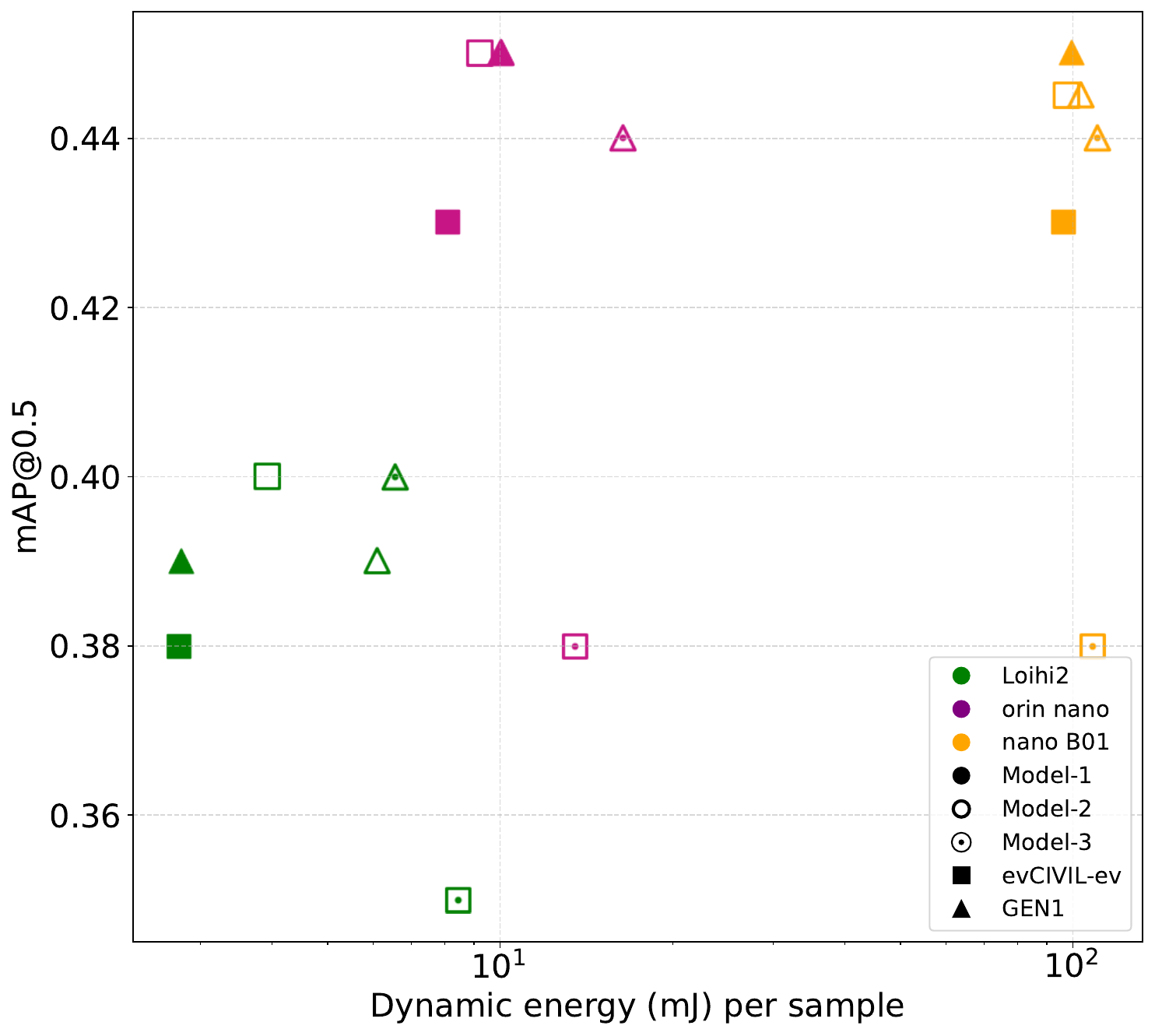}%
\label{fig:dynamic_energy_event_based}}
\caption{(a) $mAP_{0.5}$ vs.\ inference rate, and (b) $mAP_{0.5}$ vs.\ dynamic energy per sample (mJ), comparing the single-chip Loihi~2 system on the Intel Oheo Gulch platform~\cite{loihi2_oheo_gulch}, the Jetson Nano B01 edge GPU, and the Jetson Orin Nano edge GPU (batch size = 1). Results are reported for Model-1, Model-2, and Model-3 across event-based datasets.}
\label{fig_sim}
\end{figure*}

\begin{figure*}[!t]
\centering
\subfloat[]{\includegraphics[width=0.50\textwidth]{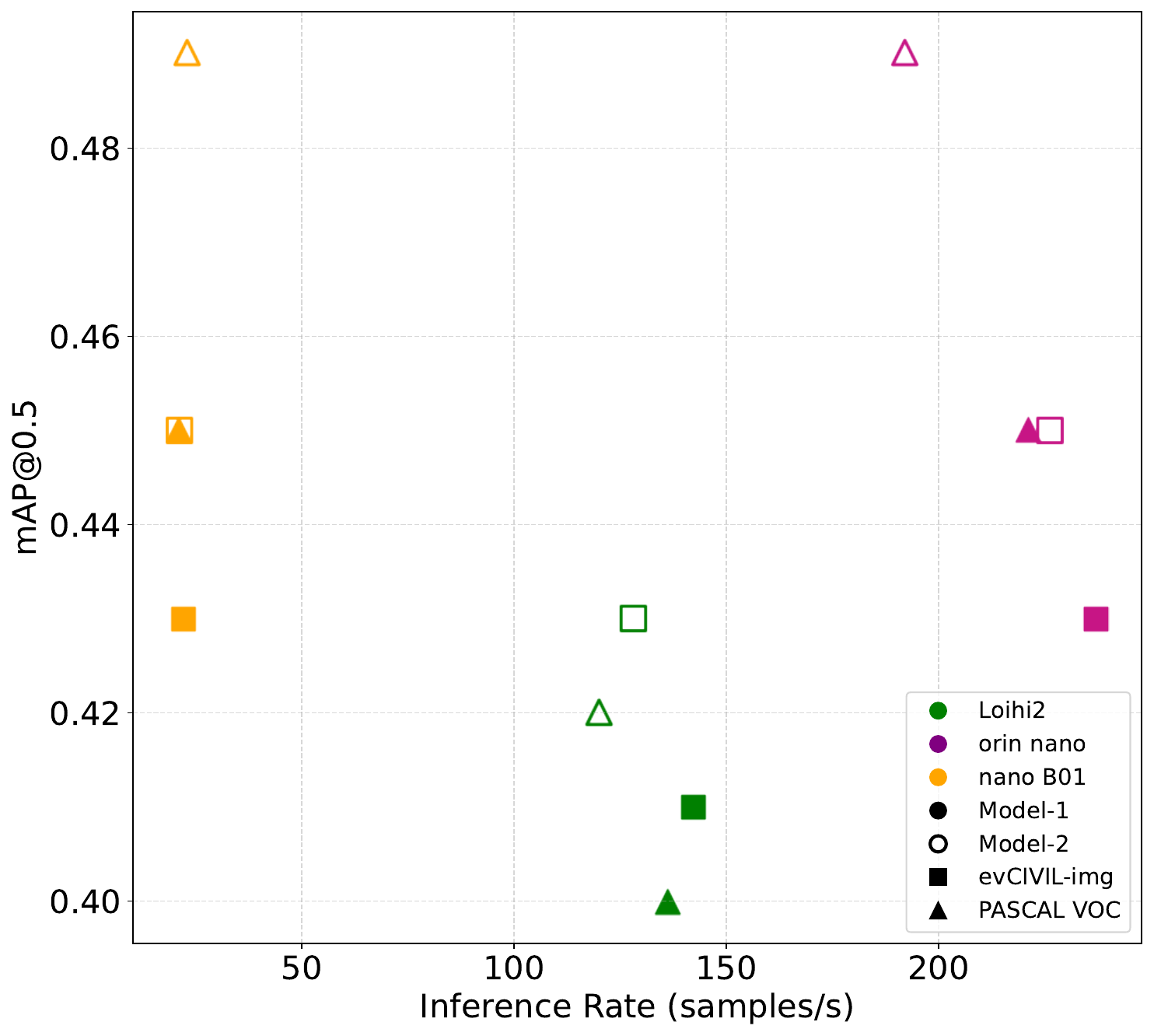}%
\label{fig:inference_rate_image_based}}
\hfil
\subfloat[]{\includegraphics[width=0.49\textwidth]{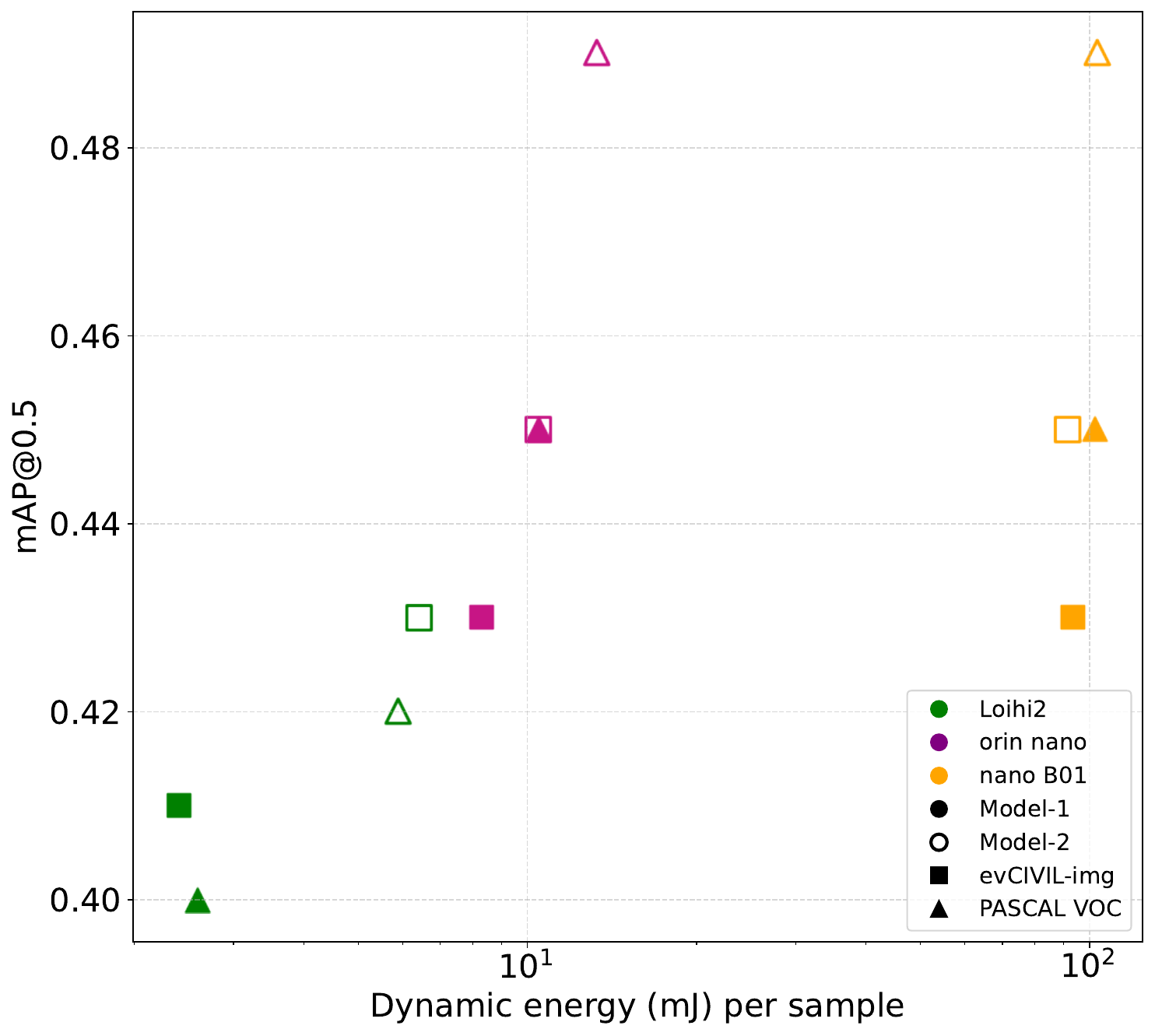}%
\label{fig:dynamic_energy_image_based}}
\caption{(a) $mAP_{0.5}$ vs.\ inference rate, and (b) $mAP_{0.5}$ vs.\ dynamic energy per sample (mJ), comparing the single-chip Loihi~2 system on the Intel Oheo Gulch platform~\cite{loihi2_oheo_gulch}, the Jetson Nano B01 edge GPU, and the Jetson Orin Nano edge GPU (batch size = 1). Results are reported for Model-1, Model-2, and Model-3 across frame-based datasets.}
\label{fig_sim}
\end{figure*}

\Cref{fig:inf_rate_event_based} and \cref{fig:inference_rate_image_based} compare the inference rates of SNNs against detection accuracy ($mAP_{0.5}$) for event-based and frame-based datasets, respectively, across all three models.
Similarly, \Cref{fig:dynamic_energy_event_based} and \cref{fig:dynamic_energy_image_based} compare dynamic energy consumption per inference against detection accuracy ($mAP_{0.5}$) for the same datasets and models per inference.
In these experiments, SNN models are executed on Loihi~2, while ANN models are deployed on the Jetson Nano B01 and Jetson Orin Nano platforms.

Across all datasets, SNNs on Loihi~2 achieve real-time inference rates ranging from 62 to 170 samples/s, which are approximately $5.8\text{--}6.95\times$ higher than those of ANNs on the Jetson Nano B01 GPU. However, ANNs on the Jetson Orin Nano outperform SNN inference rates by approximately $1.34\text{--}2.6\times$ across all models and evaluated datasets.

In terms of dynamic energy consumption, Loihi~2 is more energy-efficient than both Jetson platforms. SNNs on Loihi~2 achieve $10\text{--}35\times$ higher energy efficiency compared to ANNs on the Jetson Nano B01, and $1.61\text{--}3.74\times$ higher energy efficiency compared to ANNs on the Jetson Orin Nano.
However, SNNs in Loihi 2 shows an approximately 5–15\% reduction in $mAP_{0.5}$ compared to ANNs in Jetson Nano and MacBook CPU platforms.

\Cref{tab:EDP_table} presents the per-inference Energy–Delay Product (EDP), along with the latency (L) and total energy per inference (TE), evaluated on Loihi 2, Jetson Orin Nano, Jetson B01, and MacBook M2 CPU.  
In terms of total energy, SNN2 on Loihi 2 outperforms all other platforms, achieving at least 1.5$\times$ higher energy efficiency. Regarding EDP, the Jetson Orin Nano exhibits superior performance among the platforms due to its higher inference rate and lower latency, achieving at least 1.14$\times$ lower EDP compared to its nearest competitor, Loihi 2.  
Nonetheless, Loihi 2 maintains 1.5–10$\times$ lower EDP than the Jetson B01 and MacBook M2 CPU platforms.

\subsection{\textcolor{black}{Comparison of Power onsumption: Loihi 2 vs. NVIDIA Jetson nano and MacBook M2 CPU}}

\Cref{tab:total_power_benchmark_compare} compares the total and dynamic power consumption of Loihi 2 on the Oheo Gulch board, Jetson Orin Nano, Jetson Nano B01, and MacBook CPU.  

Being the lowest power consumer, SNN on Loihi 2 on the Oheo Gulch board outperforms ANNs running on all three other platforms. Compared to Jetson Orin Nano the nearest competitor, Loihi 2 achieves at least 2.5$\times$ higher efficiency in terms of total power consumption and at least 4.07$\times$ higher efficiency in dynamic power consumption.

\begin{table*}[t!]
\centering
\caption{Detection performance of ANN and SNN models on frame-based and event-based datasets. 
For each dataset set, the lowest value of each metric (the best value) is highlighted per model. 
All experiments on Jetson Nano devices were conducted with a batch size of 1(‘–’ indicates not applicable).}
\label{tab:EDP_table}
\fontsize{7pt}{8pt}\selectfont
\setlength{\tabcolsep}{2.5pt}
\begin{tabular*}{\textwidth}{@{\extracolsep{\fill}}ll|ccc|ccc|ccc|ccc}
\toprule
 & & \multicolumn{3}{c|}{evCIVIL -- ev} & \multicolumn{3}{c|}{evCIVIL -- fr} & \multicolumn{3}{c|}{PASCAL VOC} & \multicolumn{3}{c}{GEN1} \\
 & & TE & L & EDP & TE & L & EDP & TE & L & EDP & TE & L & EDP \\
Model & Platform & \textit{(mJ)} & \textit{(ms)} & \textit{($\mu$J)} & \textit{(mJ)} & \textit{(ms)} & \textit{($\mu$J)} & \textit{(mJ)} & \textit{(ms)} & \textit{($\mu$J)} & \textit{(mJ)} & \textit{(ms)} & \textit{($\mu$Js)} \\
\midrule
Model-1 & Loihi 2 & \textbf{13.05} & 9.37 & 122.3 & \textbf{20.06} & 14.02 & 281.2 & \textbf{15.01} & 11.03 & 165.6 & \textbf{16.50} & 11.36 & 187.5 \\
 & nano B01 & 162.39 & 43.47 & 7060.5 & 170.31 & 45.45 & 7741.7 & 184.52 & 47.61 & 8786.9 & 164.04 & 45.45 & 7456.5 \\
  & Orin Nano & 23.79 & \textbf{3.80} & \textbf{90.47} & 27.62 & \textbf{4.22} & \textbf{116.56} & 28.84 & \textbf{4.52} & \textbf{130.48} & 28.91 & \textbf{4.65} & \textbf{134.47} \\
 & CPU & 76.00 & 7.74 & 588.2 & 117.30 & 6.76 & 792.9 & 114.10 & 6.44 & 734.8 & 149.80 & 11.90 & 1782.6 \\
\hline
Model-2 & Loihi 2 & \textbf{14.02} & 9.56 & 134.0 & \textbf{19.92} & 12.60 & 250.9 & \textbf{20.30} & 13.54 & 274.9 &\textbf{25.46} & 16.75 & 426.5 \\
 & nano B01 & 159.21 & 43.48 & 6837.0 & 177.23 & 47.62 & 8440.0 & 167.48 & 43.48 & 7325.1 & 178.91 & 45.45 & 8132.2 \\
 & Orin Nano & 26.02 & \textbf{4.37} & \textbf{113.63} & 27.80 & \textbf{4.42} & \textbf{123.01} & 33.52 & \textbf{5.21} & \textbf{174.59} & 31.78 &\textbf{ 4.74} & \textbf{150.60}  \\
 & CPU & 98.00 & 10.31 & 1010.4 & 148.20 & 8.67 & 1284.9 & 138.50 & 7.82 & 1083.1 & 173.20 & 14.69 & 2544.3 \\
\hline
Model-3 & Loihi 2 & \textbf{35.00} & 24.19 & 846.8 & - & - & - & - & - & - & \textbf{28.88} & 17.65 & 509.7 \\
 & nano B01 & 188.65 & 50.00 & 9432.5 & - & - & - & - & - & - & 188.89 & 52.63 & 9941.8 \\
 & Orin Nano & 40.20 & \textbf{6.17} & \textbf{248.03} & - & - & - & - & - & - & 42.71 & \textbf{6.67} & \textbf{284.75} \\
 & CPU & 158.80 & 14.76 & 2343.9 & - & - & - & - & - & - & 242.10 & 17.76 & 4299.7 \\
\bottomrule
\multicolumn{13}{l}{\textit{TE: Total energy  (samples/s), L: Latency, EDP: Energy-Delay Product}}
\end{tabular*}
\end{table*}

\begin{table}[h!]
\centering
\caption{\textcolor{black}{Comparison of the total power and Dynamic power ($W$) in Loihi 2, CPU and nvidia-jetson across all the datasets}}
\label{tab:total_power_benchmark_compare}
\renewcommand{\arraystretch}{1.3} 
\resizebox{0.8\textwidth}{!}{
\begin{tabular}{l|c|c|c|c}
\toprule
 & Loihi 2 & nano B01 & Orin Nano & CPU\\ 
\hline \hline

Total   & 2.05 -- 2.45 & 3.5 -- 3.77 & 6.01 -- 6.71 & 9.5 -- 17.6 \\
Dynamic & 0.34 -- 0.77 & 2.06 -- 3.13 & 2.09 -- 2.55 & 9.5 -- 17.6  \\

\bottomrule
\end{tabular}%
}
\end{table}

\subsection{Benchmark results on the UAV-based tunnel inspection \\ dataset}

\begin{figure}[!t]
\centering
\includegraphics[width=1.0\textwidth]{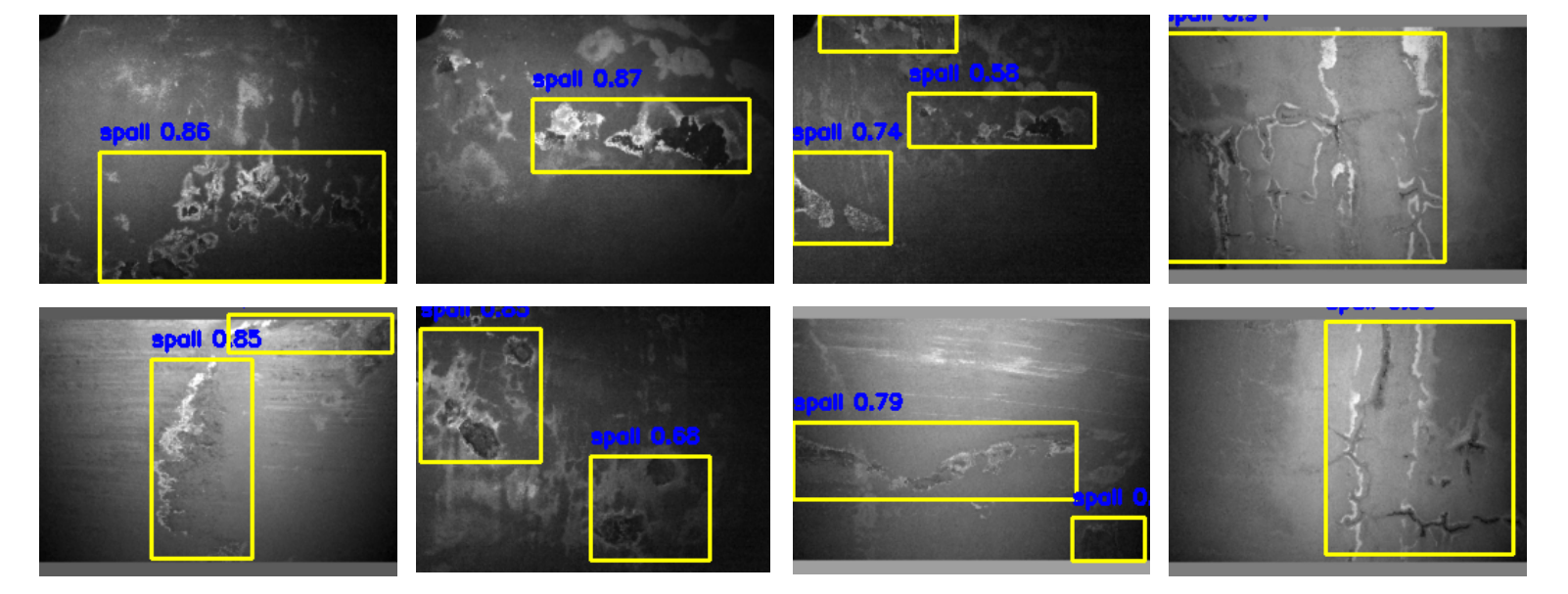}%
\caption{\textcolor{black}{Qualitative visualization of sample detections with Model-2 SNN on UAV-based tunnel dataset}}

\label{fig:qualititative_visualize_tunnel}
\end{figure}

In \cref{{tab:tunnel_benchmark}}, the detection performance of ANN and distilled SNN version of Model-2 trained and tested based on the UAV-based tunnel inspection dataset is shown.
The distilled SNN could recover 94\% of detection accuracy as of it's ANN counterpart interms of $mAP_{0.5}$ and 90\% of recovery compared to $mAP_{0.5}$.
SNN on Model-2 shows the lowest per inference dynamic energy and ANN on Jetson Orin Nano shows highest inferece rate consequently the lowest EDP.
These results are consistent with the previous evaluation results which obtained with event-based and frame-based datasets from the literature.
\Cref{fig:qualititative_visualize_tunnel} shows a qualitative visualization of detected and localized defects on test set of tunnel dataset which detected by using distilled SNN version of Model-2. It can be seens that defects are detected with high localization accuracy.

\begin{table}[htbp]
\centering
\caption{Detection performance and hardware-aware benchmarks for defect detection using Model-2 on the UAV-based Tunnel Inspection Dataset. The highest values for detection metrics and inference rate, as well as the lowest values for dynamic energy, total energy, and EDP, are highlighted.}
\label{tab:tunnel_benchmark}
\footnotesize 
\setlength{\tabcolsep}{0pt} 
\begin{tabular*}{\columnwidth}{@{\extracolsep{\fill}} l ccc ccc c}
\hline
\multirow{2}{*}{Platform} & \multicolumn{3}{c}{Detection Perf.} & Dyn. E & Rate & Total E & EDP \\
\cline{2-4}
 & $mAP_{.5}$ & $mAP_{.5:.95}$ & $F1_{iou@.5}$ & (mJ) & (Sa/s) & (mJ) & ($\mu$Js) \\
\hline
Loihi 2 & .82 & .61 & .78 & \textbf{6.2} & 122 & \textbf{20.5} & 272.86 \\
nano B01 & .87 & .63 & .82 & 97 & 23 & 172.04 & 7396 \\
Orin Nano & .87 & .63 & .82 & 12.03 & \textbf{208} & 32.41 & \textbf{155.82} \\
CPU & \textbf{.87} & \textbf{.63} & \textbf{.82} & 174.0 & 105 & 174.0 & 1642 \\
\hline
\multicolumn{8}{l}{\textit{Rate: Inference Rate (samples/s), Dyn. E: Dynamic Energy, E: Energy}}
\end{tabular*}
\end{table}

\subsection{Impact of Synaptic Operations(SOPs) on Inference Rate and Dynamic Energy Consumption on Loihi 2}

\subsubsection{Inference Rate vs. SOP}

As shown in \cref{fig:inf_vs_map}, the inference rate generally decreases as the number of SOPs increases across all three models and four datasets, regardless of model size or data modality (event-based or frame-based). For instance, compared to Prophesee GEN1, PASCAL VOC achieves a higher inference rate on Model-2, although PASCAL VOC is frame-based while GEN1 is event-based.

Interestingly, for the evCIVIL-ev dataset, Model-2, which is approximately 50\% larger than Model-1, achieves a higher inference rate (170 samples/s vs. 160 samples/s). This occurs because Model-2 exhibits a lower number of SOPs due to its higher sparsity.

Although the general trend shows that the inference rate decreases with increasing SOPs, we observed an exception when the number of SOPs is nearly identical. Specifically, Model-1 performs approximately the same number of SOPs on both evCIVIL frame-based and event-based datasets, yet achieves a higher inference rate on the event-based version. Upon further analysis, in the event-based case, only about 70\% of convolutional synapses perform at least one addition operation, leaving 30\% idle. In contrast, for the frame-based case, about 95\% of synapses are active, with only 5\% idle.
Although some of these addition operations do not result in output spikes, they still consume computational resources. Although their impact is relatively minor compared to SOP (which account only for operations that generate spikes), such effects can become noticeable when the SOP counts are nearly identical across models or datasets.

Across all datasets and all three SNN models, the inference rate in Loihi 2 ranges between 62 and 170 samples/s, confirming the feasibility of real-time inference in Loihi 2 in both event-based and frame-based detection tasks.

\subsubsection{Per-Inference Dynamic Energy Consumption vs. SOP}

\begin{figure*}[!t]
\centering
\subfloat[]{\includegraphics[width=0.49\textwidth]{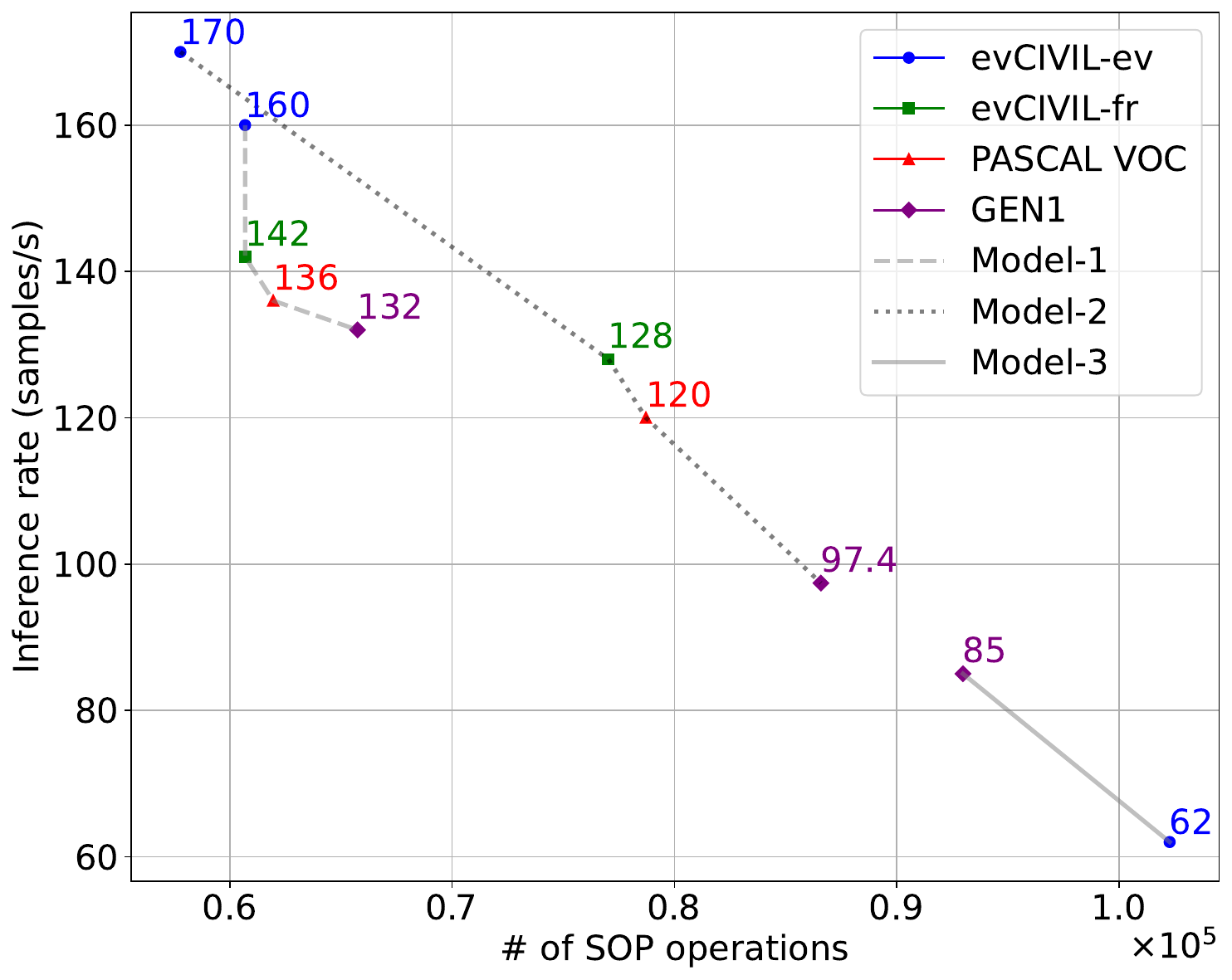}%
\label{fig:inf_vs_map}}
\hfil
\subfloat[]{\includegraphics[width=0.49\textwidth]{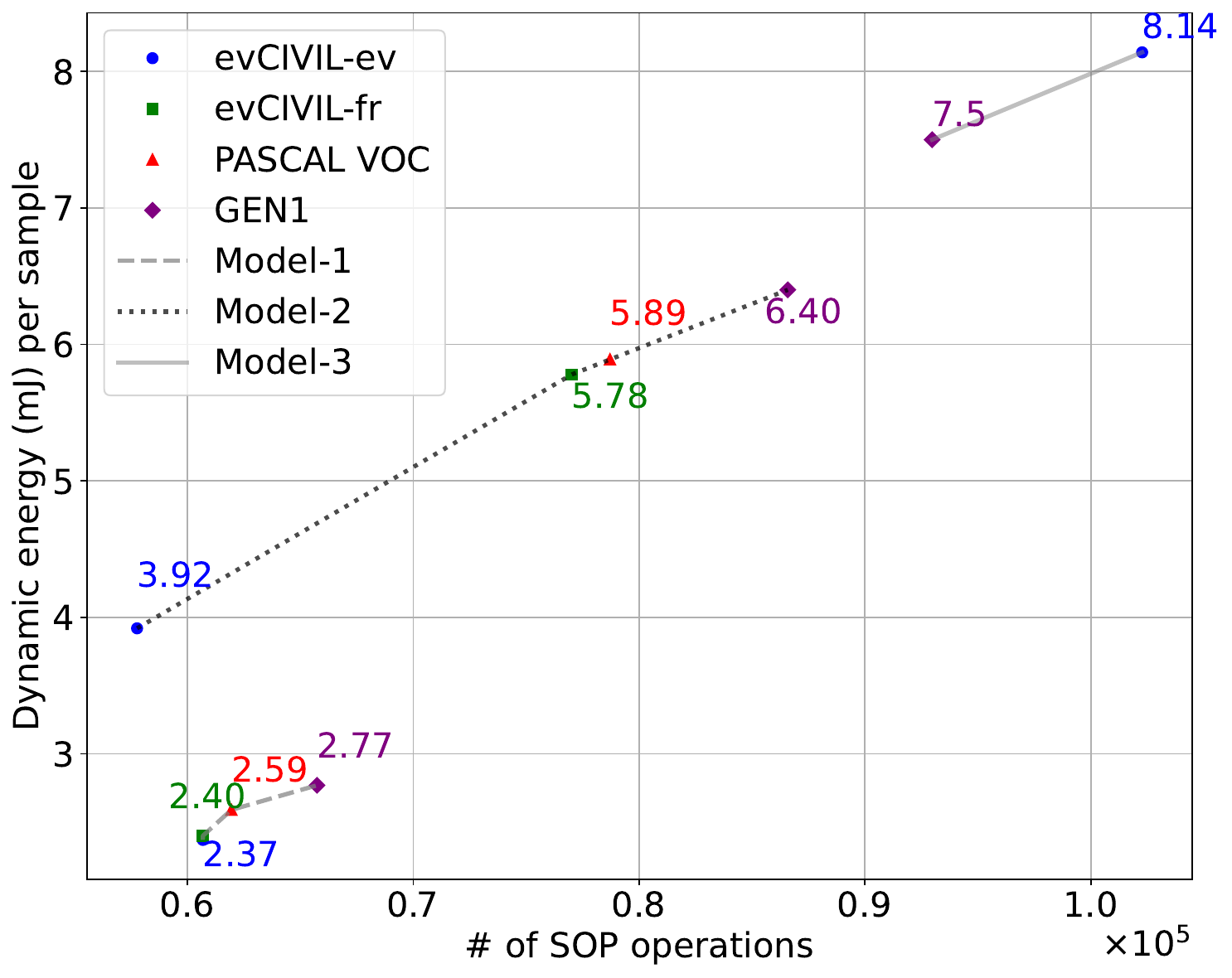}%
\label{fig:dyn_energy_vs_map}}
\caption{(a) Number of SOP operations vs. inference rate(Throughput); (b) Number of SOP operations vs. dynamic energy consumption on a single Loihi 2 chip, evaluated using four selected datasets across three models.}
\label{fig_sim}
\end{figure*}

As shown in \cref{fig:dyn_energy_vs_map}, the dynamic energy consumption per-inference generally increases with the number of SOPs across the three models and all four datasets, except for one case, that is evCIVIL-ev in Model-2. In this particular case, evCIVIL-ev in Model-2 shows higher dynamic energy consumption than in Model-1, despite having fewer SOPs. This occurs because the dynamic energy depends on both the inference rate and the dynamic power. Although Model-2 achieves a higher inference rate than Model-1, it's higher dynamic power consumption (as shown in \cref{tab:power_comparison}), leads to a higher overall dynamic energy per inference.
However, although Prophesee GEN1 consumes more power than evCIVIL-ev on Model-3, it shows lower per-inference dynamic energy, because of its higher inference rate compared to evCIVIL-ev.
Among the three models, Model-1 demonstrates the lowest dynamic power consumption, while Model-3 shows the highest overall energy consumption.

\textcolor{black}{\subsection{Summary of experimental results}}

\textcolor{black}{Our results demonstrate that ANN-SNN knowledge-distillation-aware direct SNN training enables SNNs to recover 87–100\% of the detection performance of their ANN counterparts while requiring fewer time steps, thereby reducing latency, compared to traditional ANN-to-SNN conversion methods across both frame-based and event-based datasets.} 

\textcolor{black}{In terms of power consumption, SNNs on Loihi 2 consistently exhibited lower total and dynamic power usage compared to Jetson Orin Nano, Jetson Nano B01, and MacBook M2 CPU. Across both dataset types, Loihi 2 also achieved the lowest per-inference dynamic energy consumption relative to ANNs running on other platforms.}

\textcolor{black}{With inference rates ranging from 62 to 170 samples/s, SNNs on Loihi 2 consistently demonstrated real-time inference. These rates surpass those of Jetson Nano B01, are lower than those of Jetson Orin Nano, and are comparable to ANNs on the MacBook CPU. Consequently, ANNs on Jetson Orin Nano achieved the lowest per-inference EDP among all platforms, whereas SNNs on Loihi 2 achieved the lowest per-inference total energy consumption compared to ANNs on other platforms.} 

\textcolor{black}{Evaluation on the newly introduced tunnel dataset, collected via UAV field inspections, confirmed the trends which we observed on the literature datasets. In this case, the distilled SNN retained 94\% of detection performance as of ANN counterparts in terms of $mAP_{0.5}$ while maintaining real-time inference, and exhibiting the lowest dynamic energy and power consumption on Loihi 2.}  

\textcolor{black}{Finally, we experimentally validated the impact of the number of SOP operations in SNNs on neuromorphic hardware (Loihi 2) for inference rate and per-inference dynamic energy consumption. Across frame-based and event-based data modalities, our findings reinforce the theoretical basis which is reducing the number of SOP operations increases inference rates and lowers dynamic energy. This highlights the advantage of leveraging sparsity during inference through sparsity-aware SNN training. However, reducing the number of spikes excessively can degrade detection accuracy, indicating that a careful trade-off between sparsity and performance must be maintained.}

\section{Discussion}

This study proposes three lightweight spiking neural network (SNN) architectures optimized for deployment on Intel Loihi 2, targeting real-time object detection.
We developed three lightweight SNN models that comply with both SNN design principles and the hardware constraints of Intel Loihi 2. Additionally, we present a complete pipeline covering direct SNN training, hardware-aware engineering adaptations for deployment on Loihi 2, and systematic benchmarking. The pipeline includes data preprocessing, model initialization, batch normalization mapping to Loihi hardware, distillation- and quantization-aware training, and time-step selection aligned with periodic resets.

To enhance detection performance, an ANN-to-SNN distillation framework is introduced, enabling the proposed SNN models to achieve 87–100\% of the detection accuracy of their ANN counterparts with just 8 time steps per inference on Loihi 2 which is significantly fewer time steps per inference compared to conventional ANN-to-SNN converted models which requires hundreds of time steps in object detection tasks as reported in \cite{panda2021snn} and \cite{annsnnconversion}.

Furthermore, we observed that distillation-aware direct SNN training outperforms approaches that rely on features derived from event-to-image reconstruction which highlights the effectiveness of direct temporal supervision via distillation for event-based object detection tasks using SNNs.

Benchmark comparisons demonstrate that the SNNs running on Loihi 2 outperform ANN implementations on Jetson Orin Nano, Jetson Nano B01, and a MacBook CPU in terms of power efficiency and per-inference dynamic energy consumption. Although the SNN models on Loihi 2 achieve real-time inference rates, ANN counterparts on Jetson Orin Nano  exhibit higher inference rates.

Using the PASCAL VOC dataset, \cref{tab:discussion_voc_gen1} compares our best-performing distilled SNN, Model-2, evaluated on Loihi 2 in terms of inference rate, per-inference dynamic energy, and power consumption, with the state-of-the-art SNN SpikeYOLO \cite{SpikeYOLO}, benchmarked on an RTX 3090 GPU (24 GB VRAM). The comparison also includes the ANN-based TY-v1.3-Small model \cite{gap9_tinyyolo}, deployed on GAP9 \cite{gap9_tinyyolo}, an ultra-low-power edge processor. For fairness, the GAP9 results are reported without I/O overhead to match the Loihi 2 evaluation conditions.
As per the results, although the SOTA SNN (SpikeYOLO) achieves 2.2× higher detection accuracy in terms of $mAP_{0.5:0.95}$, its inference rate on a high-end GPU is 68\% lower than that of Model-2 running on Loihi 2. In terms of model size, SpikeYOLO uses 109× more activation units/neurons than Model-2.
The detection performance of Model-2 is comparable to that of the ANN model TY-v1.3-Small, while Model-2 uses 2.2× fewer activation units. Although TY-v1.3-Small in GAP9 achieves 3.7× lower energy consumption compared to Model-2 in Loihi 2, Model-2 provides approximately 2× higher inference rate.

\begin{table*}[ht]
    \centering
    \caption{Consolidated comparison of selected benchmark results on the PASCAL VOC and GEN1 datasets, including our best SNN model alongside state-of-the-art SNNs and edge-focused ANN models. For each dataset, the lowest activation count, parameter count, and dynamic energy, as well as the highest inference rate, are highlighted.}
    \label{tab:discussion_voc_gen1}
    \fontsize{8pt}{8pt}\selectfont
    \setlength{\tabcolsep}{2.5pt}
    \begin{tabular*}{\textwidth}{@{\extracolsep{\fill}} l l l c c c c c c c @{}}
        \toprule
        Dataset & Model & Plat. & Act. & Res. & mAP & \# Act. & \# Params & Rate & Dyn. En. \\
         & & & Unit &  & 0.5:0.95 & Units(M) & (M) & (Sa/s) & (mJ) \\
        \midrule
        \multirow{3}{*}{PASCAL} & SpikeYOLO\cite{SpikeYOLO} & GPU & LIF & $640 \times 640$ & 0.64 &  84 & 41 & 38 & n/a \\
         & TY-v1.3-Small\cite{gap9_tinyyolo} & GAP9 & ReLU & $256 \times 256$ & 0.30 & 1.7 & \textbf{0.4} & 59 & \textbf{1.59} \\
         & YOLOv8n \cite{yolov8_ref} & Orin Nano & SiLU & $320 \times 320$ & 0.38 & 16.81 & 3.16 & 92 & 12.88 \\
         & Model-2 (Ours) & Loihi 2 & LIF & $224 \times 224 $ & 0.29 & \textbf{0.768} & 3.83 & \textbf{120} & 5.89 \\
        \midrule
        \multirow{3}{*}{GEN1} & SpikeYOLO\cite{SpikeYOLO} & GPU & LIF & $640 \times 640$ & 0.39 & 84 & 41 & 31 & n/a \\
         & SSD-Densenet121\cite{LOIC} & GPU & pLIF & $300 \times 300$ & 0.19 & 52 & 8.2 & 39 & n/a \\
         & YOLOv8n \cite{yolov8_ref} & Orin Nano & SiLU & $300 \times 300$ & 0.29 & 16.81 & \textbf{3.16} & 84 & 15.54 \\
         & Model-2 (Ours) & Loihi 2 & LIF & $256 \times 192$ & 0.23 & \textbf{0.768} & 3.83 & \textbf{97} & \textbf{6.40} \\
        \bottomrule
        \addlinespace
        \multicolumn{10}{l}{\footnotesize \textit{Rate: Inference Rate (samples/s), Dyn. En.: Dynamic Energy.}} \\
        \multicolumn{10}{l}{\footnotesize \textit{pLIF : parametric Leaky Integrate and Fire}}
    \end{tabular*}
\end{table*}

Using the Prophesee GEN1 dataset, \cref{tab:discussion_voc_gen1} further compares the detection performance of the SNN models SpikeYOLO \cite{SpikeYOLO} and SSD-DenseNet121-24 \cite{LOIC} with our Model-2.
Although the SOTA SNN shows 1.7× higher detection accuracy in terms of $mAP_{0.5:0.95}$, its inference rate on GPU is 68\% lower than that of Model-2 on Loihi 2. Model-2 further outperforms SSD-Densenet121-24 by 17\% in detection accuracy while being 67\% smaller in terms of activation neurons.

\textcolor{black}{Compared to lightweight, edge-focused ANNs, SNNs on Loihi 2 exhibit a considerable reduction in detection accuracy. As shown in \ref{tab:discussion_voc_gen1}, with respect to YOLOv8n on Jetson Orin Nano, our best-performing SNN model shows a 24\% reduction in detection accuracy on the PASCAL VOC dataset and a 21\% reduction on the GEN1 dataset.
However, for both datasets, SNNs on Loihi 2 achieve higher inference rates and lower dynamic energy consumption compared to YOLOv8n on Jetson Orin Nano. While the reduced detection accuracy makes deploying SNNs on Loihi 2 for safety-critical tasks potentially risky, their ability to process more frames per second can mitigate the risk of missing objects that appear in multiple frames, even though localization accuracy remains lower.}

A key factor contributing to the observed performance gap of SNNs on Intel Loihi 2 is the current hardware limitation that supports only mean-only batch normalization, rather than the standard mean-and-variance-based normalization. Next-generation Neuromorphic hardware is expected to support standard batch normalization, which will likely significantly reduce this accuracy degradation.
Moreover, anticipated support for more complex architectures—such as multi-scale branching networks and ConvLSTM modules—will better address the challenges posed by sparse event-based data.
\textcolor{black}{Additionally, rather than relying on a fixed firing threshold (e.g., 1 V), future work could explore learning optimal threshold values either globally or on a per-layer basis by leveraging concepts from parametric LIF (pLIF) neurons \cite{pLIF}. This approach may further improve accuracy, particularly if Loihi 2 evolves to support different firing thresholds for different neurons within a single model.}

\textcolor{black}{While current benchmarking excludes I/O overhead in accordance with Intel’s guidance for the Loihi 2 platform, this factor remains critical for real-world deployment scenarios such as UAV-based inspection. In such applications, sensor data transfer significantly impacts power consumption, inference rate, and overall energy efficiency.
Future iterations of Loihi that incorporate high-speed I/O interfaces will enable more comprehensive system-level evaluations, including end-to-end latency and dynamic power consumption that accounts for data transfer.
Moreover, as the neuron capacity of Loihi scales, it will support the deployment of larger and deeper spiking neural networks (SNNs), helping to further close the performance gap with ANN counterparts. For instance, Intel has recently introduced Hala Point \cite{halapoint_2024}, a 1.15-billion-neuron neuromorphic system, which is expected to become accessible for future research.}

\textcolor{black}{Intel’s Sigma-Delta Neural Networks (SDNNs) \cite{SDNN_Loihi2} represent a specialized alternative to conventional spiking neural networks (SNNs). By communicating through decimal-valued events and processing one frame per time step, SDNNs more effectively exploit temporal redundancy in video streams compared to our multi-timestep SNN approach. Reported SDNN benchmarks on an eight-chip Loihi 2 cluster demonstrate an energy consumption of approximately 8 mJ and a throughput of 314 samples/s—roughly twice the speed of our implementation.
However, the operating principles of SDNNs differ fundamentally from those of standard SNNs, and they have not yet demonstrated performance surpassing that of artificial neural networks (ANNs). Consequently, this work does not explore SDNN models, as they are tightly coupled to the Loihi 2 architecture rather than representing a generalizable approach like SNNs, which are more broadly applicable across neuromorphic platforms.}

Our evaluation on the ev-CIVIL dataset \cite{Gamage2025_evCIVIL} demonstrates that event-based detection consistently outperforms frame-based methods under challenging lighting conditions, such as low-light and overexposed scenes, while maintaining energy efficiency and real-time inference rates. These results highlight the potential of integrating DVS cameras with Loihi 2 for in-situ, real-time inspection applications. However, the reliance of DVS sensors on relative motion introduces directionality constraints, and the absence of color information may lead to missed or ambiguous anomalies.
Consequently, a hybrid sensing approach that leverages both frame- and event-based modalities represents a promising direction for future work. By fusing the high-dynamic-range temporal precision of event cameras with the static spatial structure and color information provided by frame-based sensors, overall detection robustness can be significantly enhanced for complex structural inspection scenarios.

Although SNNs running on neuromorphic processors such as Loihi 2 \cite{brainchip_akida} and Akida \cite{brainchip_akida} represent a promising path toward energy-efficient deep learning, these chips still require substantial development in terms of computational capacity, energy efficiency, and architectural support for more complex models necessary for high-accuracy detection as previously mentioned. At the same time, rapid progress is being made in improving the energy efficiency of ANNs for edge devices through techniques such as pruning \cite{mit_quantization_lecture2020}, quantization \cite{mit_quantization_lecture2020}, and neural architecture search \cite{MIT_NAS_Lecture07}. Energy-efficient platforms including GPUs like the Jetson Orin Nano and Xavier \cite{lightweight_edge_devices}, as well as ultra-low-power microcontrollers with neural accelerators such as GreenWaves’ GAP9 \cite{gap9_tinyyolo} are gaining significant commercial momentum.
Therefore, future neuromorphic research needs to keep up with advances in energy-optimized ANNs and emerging accelerators. However, we believe that innovation in both directions will drive important breakthroughs in energy-efficient real-time detection with improved accuracy.

\section{Conclusion}

In this work, we addressed real-time event-based and frame-based object detection covering both regularly shaped and irregularly shaped objects using Spiking Neural Networks (SNNs) which are designed to fit in state-of-the-art neuromorphic hardware called intel Loihi 2.

We developed three lightweight SNN architectures that comply with both SNN design principles and the hardware constraints of Intel Loihi 2. Additionally, we present a complete pipeline covering direct SNN training, hardware-aware engineering adaptations for deployment on Loihi 2, and systematic benchmarking.

Comparative benchmarking shows that directly trained SNNs achieve 11–27\% lower detection accuracy than ANNs. However, ANN-to-SNN knowledge distillation allows SNNs to retain 87–100\% of the original accuracy of ANNs while performing inference while maintaining a less number of inference time steps compared to as of ANN to SNN converted models.
Also, Distillation-aware direct SNN training for object detection task outperforms approaches which atr based on feature maps from event-to-image reconstruction architectures.

We also demonstrate that increasing synaptic operations (SOPs),  raises dynamic energy consumption and reduces inference rates, regardless of event- or frame-based input. Power profiling on Loihi 2 reveals that static power dominates dynamic power, ranging from 2.5× to 5.2× higher across all models.

SNNs on Loihi 2 outperform ANN implementations on Jetson Nano B01 and Apple M2 CPU in terms of inference rate, dynamic energy, and total power. 
Compared to the Jetson Orin Nano, Loihi 2 achieves at least 2.5× higher efficiency in total power and 4.07× higher efficiency in dynamic power. 
Also, SNNs on Loihi 2 shows 1.61–3.74× higher efficiency interms of per inference dynamic energy consumption than ANNs on Jetson Orin Nano, 
Though SNNs on Loihi 2 shows real-time inference rates range from 62 to 170 samples/s, ANNs on Orin Nano maintain 1.34–2.6× higher inference rates as of SNNs on Loihi 2.

Results of the ealuations of our newly introduced UAV-based tunnel inspection dataset on considered hardware plarforms were consistent with the results of our developed models which were evaluated across two event-based and frameb-based data sets from the literature. Howerver, SNNs currently lag behind state-of-the-art ANNs like YOLOv8n, majorly in terms of detection accuracy, which would be further enchanced by advancing the neuromorphic hardware to facilitate for advanced SNN architectures such as Conv-LSTMs and network branching.

Despite current limitations, our results demonstrate that neuromorphic systems can perform real-time object detection while maintaining high energy efficiency. Ultimately, this work establishes a scalable baseline for the future deployment of real-time, energy-efficient, and autonomous intelligence at the edge.

\section*{Data and Code Availability}

The source code and UAV tunnel data are available at

\url{https://github.com/gwgknudayanga/Realtime-frame--and-event-based-detection-with-SNN-on-Neuromorphic-hardware} and \\
\url{https://drive.google.com/drive/folders/1ftKEMairrUScP5KXW6pQDhwqanxOuwxD} respectively.

\section*{Funding}
This project has received funding from the European Union’s Horizon 2020 Research and Innovation Programme under the Marie Sklodowska-Curie Grant Agreement No. 953454.

\section*{Acknowledgments}

Special Thanks goes to Intel Cooperation for providing Loihi 2 access for the benchmarking.

\section*{Generative AI usage}

During the preparation of this work the authors used ChatGPT (OpenAI) in order to correct grammar and spelling and improve the clarity of the writing. After using this tool/service, the authors reviewed and edited the content as needed and take full responsibility for the content of the published article.

\bibliographystyle{elsarticle-num}
\bibliography{bibliography}

@article{autonomous_uav,
title = {Spiking Neural Networks for event-based action recognition: A new task to understand their advantage},
journal = {Neurocomputing},
volume = {611},
pages = {128657},
year = {2025},
issn = {0925-2312},
doi = {https://doi.org/10.1016/j.neucom.2024.128657},
author = {Alex Vicente-Sola and Davide L. Manna and Paul Kirkland and Gaetano Di Caterina and Trevor J. Bihl},
}

@article{autonomous_drive,
title = {Real-time Object Detection in Autonomous Vehicles with YOLO},
journal = {Procedia Computer Science},
volume = {246},
pages = {2792-2801},
year = {2024},
note = {28th International Conference on Knowledge Based and Intelligent information and Engineering Systems (KES 2024)},
issn = {1877-0509},
doi = {https://doi.org/10.1016/j.procs.2024.09.392},
url = {https://www.sciencedirect.com/science/article/pii/S1877050924024293},
author = {Nusaybah M. Alahdal and Felwa Abukhodair and Leila Haj Meftah and Asma Cherif}
}

@Article{structural_inspection_review,
AUTHOR = {Luo, Kui and Kong, Xuan and Zhang, Jie and Hu, Jiexuan and Li, Jinzhao and Tang, Hao},
TITLE = {Computer Vision-Based Bridge Inspection and Monitoring: A Review},
JOURNAL = {Sensors},
VOLUME = {23},
YEAR = {2023},
NUMBER = {18},
ARTICLE-NUMBER = {7863},
PubMedID = {37765920},
ISSN = {1424-8220},
DOI = {10.3390/s23187863}
}

@article{low_light_img_review,
title = {A comprehensive experiment-based review of low-light image enhancement methods and benchmarking low-light image quality assessment},
journal = {Signal Processing},
volume = {204},
pages = {108821},
year = {2023},
issn = {0165-1684},
doi = {https://doi.org/10.1016/j.sigpro.2022.108821},
author = {Muhammad Tahir Rasheed and Daming Shi and Hufsa Khan}
}

@ARTICLE{event_vision_survey,
  author={Gallego, Guillermo and Delbrück, Tobi and Orchard, Garrick and Bartolozzi, Chiara and Taba, Brian and Censi, Andrea and Leutenegger, Stefan and Davison, Andrew J. and Conradt, Jörg and Daniilidis, Kostas and Scaramuzza, Davide},
  journal={IEEE Transactions on Pattern Analysis and Machine Intelligence}, 
  title={Event-Based Vision: A Survey}, 
  year={2022},
  volume={44},
  number={1},
  pages={154-180},
  doi={10.1109/TPAMI.2020.3008413}}

@article{YOLOv3_ref,
  title={YOLOv3: An Incremental Improvement},
  author={Redmon, Joseph and Farhadi, Ali},
  journal={arXiv preprint arXiv:1804.02767},
  year={2018}
}

@InProceedings{SSD_ref,
author="Liu, Wei
and Anguelov, Dragomir
and Erhan, Dumitru
and Szegedy, Christian
and Reed, Scott
and Fu, Cheng-Yang
and Berg, Alexander C.",
editor="Leibe, Bastian
and Matas, Jiri
and Sebe, Nicu
and Welling, Max",
title="SSD: Single Shot MultiBox Detector",
booktitle="Computer Vision -- ECCV 2016",
year="2016",
publisher="Springer International Publishing",
address="Cham",
pages="21--37",
}

@Article{resource_limit_review,
AUTHOR = {Ngo, Dat and Park, Hyun-Cheol and Kang, Bongsoon},
TITLE = {Edge Intelligence: A Review of Deep Neural Network Inference in Resource-Limited Environments},
JOURNAL = {Electronics},
VOLUME = {14},
YEAR = {2025},
NUMBER = {12},
ARTICLE-NUMBER = {2495},
ISSN = {2079-9292}
}

@article{lightweight_edge_devices,
  author  = {Mittal, P.},
  title   = {A comprehensive survey of deep learning-based lightweight object detection models for edge devices},
  journal = {Artificial Intelligence Review},
  volume  = {57},
  number  = {9},
  pages   = {242},
  year    = {2024},
  doi     = {10.1007/s10462-024-10877-1}
}

@article{edge_device_comparison,
  author  = {Chen, Jiongzhou and Ran, Xukan},
  title   = {Deep Learning With Edge Computing: A Review},
  journal = {Proceedings of the IEEE},
  volume  = {107},
  number  = {8},
  pages   = {1655--1674},
  year    = {2019},
  doi     = {10.1109/JPROC.2019.2921977}
}

@inproceedings{edgeYOLO2023,
  author    = {Liu, Shuailei and Zha, Junkai and Sun, Jie and Li, Zhilin and Wang, Gang},
  booktitle = {Proceedings of the 2023 42nd Chinese Control Conference (CCC)}, 
  title     = {EdgeYOLO: An Edge-Real-Time Object Detector}, 
  year      = {2023},
  pages     = {7507--7512},
  address   = {Tianjin, China},
  doi       = {10.23919/CCC58697.2023.10239786}
}

@inproceedings{SSD_MobileNet2025,
  title={MobileNetV2: Inverted Residuals and Linear Bottlenecks},
  author={Sandler, Mark and Howard, Andrew and Zhu, Menglong and Zhmoginov, Andrey and Chen, Liang-Chieh},
  booktitle={Proceedings of the IEEE/CVF Conference on Computer Vision and Pattern Recognition (CVPR)},
  pages={4510--4520},
  year={2018}
}

@article{neuro_drone_journal,
  author  = {Paredes-Vall{\'e}s, Federico and Hagenaars, Jesse J. and Dupeyroux, Julien and Stroobants, Stein and Xu, Yingfu and de Croon, Guido C. H. E.},
  title   = {Fully neuromorphic vision and control for autonomous drone flight},
  journal = {Science Robotics},
  volume  = {9},
  number  = {90},
  pages   = {eadi0591},
  year    = {2024},
  month   = {May},
  doi     = {10.1126/scirobotics.adi0591},
  note    = {Erratum: \textit{Science Robotics}, vol. 9, no. 91, p. eadr0223, Jun 2024}
}

@article{neuro_drone_obstacle,
  title={Real-Time Neuromorphic Navigation: Integrating Event-Based Vision and Physics-Driven Planning on a Parrot Bebop2 Quadrotor},
  author={Amogh Joshi and Sourav Sanyal and Kaushik Roy},
  journal={ArXiv},
  year={2024},
  volume={abs/2407.00931},
  url={https://api.semanticscholar.org/CorpusID:270870393}
}

@article{Roy2019,
  author  = {Roy, Kaushik and Jaiswal, Akhilesh and Panda, Priyadarshini},
  title   = {Towards spike-based machine intelligence with neuromorphic computing},
  journal = {Nature},
  year    = {2019},
  volume  = {575},
  number  = {7784},
  pages   = {607--617},
  doi     = {10.1038/s41586-019-1677-2},
}

@inproceedings{v2e_paper,
  author    = {Hu, Yuhuang and Liu, Shih-Chii and Delbruck, Tobi},
  title     = {v2e: From Video Frames to Realistic DVS Events},
  booktitle = {Proceedings of the 2021 IEEE/CVF Conference on Computer Vision and Pattern Recognition Workshops (CVPRW)},
  year      = {2021},
  pages     = {1312--1321},
  address   = {Virtual},
  doi       = {10.1109/CVPRW53098.2021.00144}
}

@inproceedings{SpikeYOLO,
  author    = {Luo, Xinhao and Yao, Man and Chou, Yuhong and Xu, Bo and Li, Guoqi},
  title     = {Integer-Valued Training and Spike-Driven Inference Spiking Neural Network for High-Performance and Energy-Efficient Object Detection},
  booktitle = {Proceedings of the European Conference on Computer Vision (ECCV)},
  year      = {2024},
  note      = {Best Paper Candidate}
}

@inproceedings{SpikingYOLO,
  author    = {Kim, Seijoon and Park, Seongsik and Na, Byunggook and Yoon, Sungroh},
  title     = {Spiking-YOLO: Spiking Neural Network for Energy-Efficient Object Detection},
  booktitle = {Proceedings of the AAAI Conference on Artificial Intelligence},
  volume    = {34},
  number    = {07},
  pages     = {11270--11277},
  year      = {2020},
  doi       = {10.1609/aaai.v34i07.6787}
}

@inproceedings{LOIC,
  author    = {Cordone, Lo{\"i}c and Miramond, Beno{\^i}t and Thierion, Philippe},
  title     = {Object Detection with Spiking Neural Networks on Automotive Event Data},
  booktitle = {2022 International Joint Conference on Neural Networks (IJCNN)}, 
  year      = {2022},
  pages     = {1--8},
  address   = {Padua, Italy},
  doi       = {10.1109/IJCNN55064.2022.9892618}
}

@InProceedings{hybrid_ANN_SNN1,
author="Kugele, Alexander
and Pfeil, Thomas
and Pfeiffer, Michael
and Chicca, Elisabetta",
editor="Bauckhage, Christian
and Gall, Juergen
and Schwing, Alexander",
title="Hybrid SNN-ANN: Energy-Efficient Classification and Object Detection for Event-Based Vision",
booktitle="Pattern Recognition",
year="2021",
publisher="Springer International Publishing",
address="Cham",
pages="297--312",
isbn="978-3-030-92659-5"
}

@InProceedings{Recur_SNN,
author="Wang, Ziming
and Wang, Ziling
and Li, Huaning
and Qin, Lang
and Jiang, Runhao
and Ma, De
and Tang, Huajin",
editor="Leonardis, Ale{\v{s}}
and Ricci, Elisa
and Roth, Stefan
and Russakovsky, Olga
and Sattler, Torsten
and Varol, G{\"u}l",
title="EAS-SNN: End-to-End Adaptive Sampling and Representation for Event-Based Detection with Recurrent Spiking Neural Networks",
booktitle="Computer Vision -- ECCV 2024",
year="2025",
publisher="Springer Nature Switzerland",
address="Cham",
pages="310--328",
isbn="978-3-031-73027-6"
}

@article{SDNN_Loihi2,
  title   = {Sigma-Delta Neural Network Conversion on Loihi 2},
  author  = {Brehove, Matthew and Tumpa, Sadia Anjum and Kyubwa, Espoir and Menon, Naresh and Vijaykrishnan, N.},
  journal = {arXiv preprint arXiv:2505.06417},
  year    = {2025},
}

@inproceedings{CarSNN,
  author    = {Viale, A. and Marchisio, A. and Martina, M. and Masera, G. and Shafique, M.},
  title     = {{CarSNN}: An Efficient Spiking Neural Network for Event-Based Autonomous Cars on the Loihi Neuromorphic Research Processor},
  booktitle = {2021 International Joint Conference on Neural Networks (IJCNN)},
  year      = {2021},
  pages     = {1--10},
  address   = {Virtual (Shenzhen, China)},
  doi       = {10.1109/IJCNN52387.2021.9533738}
}

@software{yolov8_ref,
  author = {Jocher, Glenn and Chaurasia, Ayush and Qiu, Jing},
  title  = {{YOLOv8}: Ultralytics real-time object detection},
  year   = {2023},
  url    = {https://github.com/ultralytics/ultralytics}
}

@article{li2022yolov6,
  author  = {Li, Chuyi and Li, Lulu and Jiang, Humphrey and Weng, Kai and Geng, Yifei and Li, Liang and Ke, Zaidan and Li, Qingyuan and Cheng, Meng and Nie, Weiqiang and others},
  title   = {{YOLOv6}: A Single-Stage Object Detection Framework for Industrial Applications},
  journal = {arXiv preprint arXiv:2209.02976},
  year    = {2022}
}

@misc{loihi_profile_tute,
  author       = {lava-nc contributors},
  title        = {lava-dl: {Deep Learning Library for Lava} — {PilotNet SNN Benchmark Notebook}},
  howpublished = {GitHub repository},
  year         = {2025},
  url          = {https://github.com/lava-nc/lava-dl/blob/main/tutorials/lava/lib/dl/netx/pilotnet_snn/benchmark.ipynb},
  note         = {Accessed: 26-Nov-2025}
}

@inproceedings{hats,
  author    = {Sironi, Amos and Brambilla, Manuele and Bourdis, Nicolas and Lagorce, Xavier and Benosman, Ryad},
  title     = {{HATS}: Histograms of Averaged Time Surfaces for Robust Event-Based Object Classification},
  booktitle = {Proceedings of the IEEE Conference on Computer Vision and Pattern Recognition (CVPR)},
  year      = {2018},
  pages     = {1731--1740},
  address   = {Salt Lake City, UT, USA},
  doi       = {10.1109/CVPR.2018.00186}
}

@inproceedings{voxelgrid,
  author    = {Gehrig, Daniel and Loquercio, Antonio and Derpanis, Konstantinos G. and Scaramuzza, Davide},
  title     = {End-to-End Learning of Representations for Asynchronous Event-Based Data},
  booktitle = {Proceedings of the IEEE/CVF International Conference on Computer Vision (ICCV)},
  year      = {2019},
  pages     = {5633--5643},
  address   = {Seoul, South Korea},
  doi       = {10.1109/ICCV.2019.00573}
}

@article{snn_survey,
  author  = {Nunes, J. D. and Carvalho, M. and Carneiro, D. and Cardoso, J. S.},
  title   = {Spiking Neural Networks: A Survey},
  journal = {IEEE Access},
  volume  = {10},
  pages   = {60738--60764},
  year    = {2022},
  doi     = {10.1109/ACCESS.2022.3179968}
}

@inproceedings{pLIF,
  author    = {Fang, Wei and Yu, Zhaofei and Chen, Yanqi and Masquelier, Timoth{\'e}e and Huang, Tiejun and Tian, Yonghong},
  title     = {Incorporating Learnable Membrane Time Constant to Enhance Learning of Spiking Neural Networks},
  booktitle = {Proceedings of the IEEE/CVF International Conference on Computer Vision (ICCV)},
  year      = {2021},
  pages     = {2641--2651},
  doi       = {10.1109/ICCV48922.2021.00265}
}

@article{snn_backprop,
  author  = {Dampfhoffer, M. and Mesquida, T. and Valentian, A. and Anghel, L.},
  title   = {Backpropagation-Based Learning Techniques for Deep Spiking Neural Networks: A Survey},
  journal = {IEEE Transactions on Neural Networks and Learning Systems},
  volume  = {35},
  number  = {9},
  pages   = {11906--11921},
  year    = {2024},
  doi     = {10.1109/TNNLS.2023.3263008}
}

@inproceedings{slayer,
  author    = {Shrestha, Sumit Bam and Orchard, Garrick},
  title     = {{SLAYER}: Spike Layer Error Reassignment in Time},
  booktitle = {Advances in Neural Information Processing Systems (NeurIPS)},
  volume    = {31},
  year      = {2018}
}

@article{annsnnconversion,
  author  = {Roy, Kaushik and Jaiswal, Akhilesh R. and Panda, Priyadarshini},
  title   = {Towards spike-based machine intelligence with neuromorphic computing},
  journal = {Nature},
  volume  = {575},
  number  = {7784},
  pages   = {607--617},
  year    = {2019},
  doi     = {10.1038/s41586-019-1677-2}
}

@misc{panda2021snn,
  author       = {Panda, Priyadarshini},
  title        = {Spiking Neural Networks},
  howpublished = {{ESWEEK} 2021 Tutorial, Yale University},
  year         = {2021},
  url          = {https://www.youtube.com/watch?v=7TybETlCslM},
  note         = {Accessed: 2025-04-28}
}

@techreport{intel_loihi2,
  author      = {{Intel Corporation}},
  title       = {Neuromorphic Computing: {Loihi 2} Technology Brief},
  institution = {Intel Corporation},
  year        = {2025},
  url         = {https://www.intel.com/content/www/us/en/research/neuromorphic-computing-loihi-2-technology-brief.html},
  note        = {Accessed: 2025-05-01}
}

@misc{lava_framework_video,
  author       = {Wild, A. and Richter, M.},
  title        = {Lava: Open-Source Framework for Neuro-Inspired Applications},
  howpublished = {Online Video},
  year         = {2023},
  url          = {https://www.youtube.com/watch?v=vXZukQ6A79k},
  note         = {Accessed: 2025-05-01}
}

@misc{lava_dl,
  author       = {{Lava Neuromorphic Computing Community}},
  title        = {lava-dl: Deep Learning Library for Lava},
  howpublished = {GitHub Repository},
  year         = {2025},
  url          = {https://github.com/lava-nc/lava-dl},
  note         = {Accessed: 2025-05-01}
}

@misc{ibm_neuromorphic_computing,
  author       = {{IBM}},
  title        = {What Is Neuromorphic Computing?},
  year         = {2025},
  url          = {https://www.ibm.com/think/topics/neuromorphic-computing},
  note         = {Accessed: 2025-05-01}
}

@misc{brainchip_akida,
  author       = {{BrainChip Inc.}},
  title        = {Akida --- {BrainChip} Neuromorphic Chip Overview},
  year         = {2025},
  url          = {https://brainchip.com/neuromorphic-chip-maker-takes-aim-at-the-edge},
  note         = {Accessed: Dec. 7, 2025}
}

@misc{synsense_dynap_cnn,
  author       = {{SynSense}},
  title        = {{DYNAP-CNN} --- Event-Driven Neuromorphic {AI} Processor for Vision Processing},
  year         = {2025},
  url          = {https://open-neuromorphic.org/neuromorphic-computing/hardware},
  note         = {Accessed: Dec. 7, 2025}
}

@article{furber2014spinnaker,
  author  = {Furber, Steve and Galluppi, Francesco and Temple, Steve and Plana, Luis A.},
  title   = {{SpiNNaker}: A Spiking Neural Network Architecture},
  journal = {Journal of Neural Engineering},
  volume  = {11},
  number  = {5},
  pages   = {056021},
  year    = {2014},
  publisher = {IOP Publishing}
}

@misc{intel_loihi,
  author       = {{Intel Corporation}},
  title        = {Loihi Neuromorphic Research Chip},
  year         = {2018},
  url          = {https://open-neuromorphic.org/neuromorphic-computing/hardware},
  note         = {Accessed: Dec. 7, 2025}
}

@misc{mit_quantization_lecture2020,
  author       = {{MIT 6.S965}},
  title        = {Lecture 05 - Quantization ({Part I})},
  year         = {2020},
  url          = {https://www.youtube.com/watch?v=AlASZb93rrc},
  note         = {Accessed: May 1, 2025}
}

@misc{mit2020knowledge,
  author       = {{MIT HAN Lab}},
  title        = {Lecture 10 - Knowledge Distillation | {MIT 6.S965}},
  year         = {2020},
  url          = {https://www.youtube.com/watch?v=tT9Lnt6stwA},
  note         = {Accessed: May 1, 2025}
}

@article{pascal_voc,
  author  = {Everingham, Mark and Van Gool, Luc and Williams, Christopher K. I. and Winn, John and Zisserman, Andrew},
  title   = {The {PASCAL} Visual Object Classes ({VOC}) Challenge},
  journal = {International Journal of Computer Vision},
  volume  = {88},
  number  = {2},
  pages   = {303--338},
  year    = {2010},
  publisher = {Springer}
}

@inproceedings{gen1,
  author    = {de Tournemire, Pierre and Nitti, Davide and Perot, Etienne and Migliore, Davide and Sironi, Amos},
  title     = {A Large Scale Event-Based Detection Dataset for Automotive},
  booktitle = {Proceedings of the IEEE/CVF Conference on Computer Vision and Pattern Recognition (CVPR) Workshops},
  year      = {2020},
  pages     = {346--347},
  doi       = {10.1109/CVPRW50498.2020.00173}
}

@inproceedings{gap9_tinyyolo,
  author    = {Moosmann, Julian and Bonazzi, Pietro and Li, Yawei and Bian, Sizhen and Mayer, Philipp and Benini, Luca and Magno, Michele},
  title     = {{Ultra-Efficient On-Device Object Detection on AI-Integrated Smart Glasses with TinyissimoYOLO}},
  booktitle = {Proceedings of the IEEE International Conference on Acoustics, Speech and Signal Processing (ICASSP)},
  year      = {2024},
  doi       = {10.1109/ICASSP48485.2024.10446059}
}

@inproceedings{repvgg,
  author    = {Ding, Xiaohan and Zhang, Xiangyu and Ma, Ningning and Han, Jungong and Ding, Guiguang and Sun, Jian},
  title     = {{RepVGG}: Making {VGG}-Style {ConvNets} Great Again},
  booktitle = {Proceedings of the IEEE/CVF Conference on Computer Vision and Pattern Recognition (CVPR)},
  year      = {2021},
  pages     = {13733--13742}
}

@article{anchorfree_pros,
  author  = {Kong, Tao and Sun, Fuchun and Liu, Huaping and Jiang, Yuning and Li, Lei and Shi, Jianbo},
  title   = {{FoveaBox}: {Beyond} Anchor-Based Object Detector},
  journal = {IEEE Transactions on Image Processing},
  volume  = {29},
  pages   = {7389--7398},
  year    = {2020},
  doi     = {10.1109/TIP.2020.3002345}
}

@article{snn_constraints,
  author  = {Rueckauer, B. and Wen, Y. and Liu, K. and Shi, Y. and Cauwenberghs, G.},
  title   = {Going Deeper in Spiking Neural Networks: {VGG} and Residual Architectures},
  journal = {Frontiers in Neuroscience},
  volume  = {13},
  pages   = {95},
  year    = {2019}
}

@misc{fang2020spikingjelly,
  author       = {Fang, Wei and Chen, Yanqi and Ding, Jianhao and Chen, Ding and Yu, Zhaofei and Masquelier, Timoth{\'e}e and Chen, Yonghong and Huang, Tiejun and Tian, Yonghong},
  title        = {{SpikingJelly}},
  howpublished = {\url{https://github.com/fangwei123456/spikingjelly}},
  year         = {2020},
  note         = {Accessed: May 28, 2025}
}

@misc{inrc_deep_learning_loihi2,
  author       = {{Intel Neuromorphic Research Community}},
  title        = {Deep Learning Applications for {Loihi 2}},
  howpublished = {\url{https://intel-ncl.atlassian.net/wiki/spaces/INRC/pages/1969225744/}},
  year         = {2023},
  note         = {Accessed: May 1, 2025}
}

@inproceedings{pytorch,
  author    = {Paszke, Adam and Gross, Sam and Massa, Francisco and Lerer, Adam and Bradbury, James and Chanan, Gregory and Killeen, Trevor and Lin, Zeming and Gimelshein, Natalia and Antiga, Luca and others},
  title     = {{PyTorch}: An Imperative Style, High-Performance Deep Learning Library},
  booktitle = {Advances in Neural Information Processing Systems (NeurIPS)},
  year      = {2019}
}

@inproceedings{batchnorm,
  author    = {Ioffe, Sergey and Szegedy, Christian},
  title     = {Batch Normalization: Accelerating Deep Network Training by Reducing Internal Covariate Shift},
  booktitle = {Proceedings of the 32nd International Conference on Machine Learning (ICML)},
  year      = {2015},
  pages     = {448--456}
}

@article{rebecq,
  author  = {Rebecq, Henri and Ranftl, Ren{\'e} and Koltun, Vladlen and Scaramuzza, Davide},
  title   = {High Speed and High {HDR} Video with an Event Camera},
  journal = {IEEE Transactions on Pattern Analysis and Machine Intelligence (TPAMI)},
  volume  = {43},
  number  = {6},
  pages   = {1964--1980},
  year    = {2021}
}

@inproceedings{teed2020raft,
  author    = {Teed, Zachary and Deng, Jia},
  title     = {{RAFT}: Recurrent All-Pairs Field Transforms for Optical Flow},
  booktitle = {Proceedings of the European Conference on Computer Vision (ECCV)},
  year      = {2020},
  pages     = {402--419},
  address   = {Glasgow, UK (Virtual)}
}

@misc{loihi2_oheo_gulch,
  author       = {{Intel Corporation}},
  title        = {Intel Advances Neuromorphic with {Loihi 2}, New {Lava} Software Framework and New Partners},
  howpublished = {Intel Newsroom Press Release},
  year         = {2021},
  url          = {https://www.intel.com/content/www/us/en/newsroom/news/intel-advances-neuromorphic-loihi-2-lava.html},
  note         = {Accessed: 2025-12-08}
}

@misc{lava_dl_pilotnet_sdnn,
  author       = {lava-nc contributors},
  title        = {lava-dl: {PilotNet SDNN} inference tutorial},
  year         = {2025},
  url          = {https://github.com/lava-nc/lava-dl/tree/main/tutorials/lava/lib/dl/netx/pilotnet_sdnn},
  note         = {Accessed: Dec. 7, 2025}
}

@misc{yahboom_jetson_nano,
  author       = {{Yahboom}},
  title        = {Jetson Nano 4GB B01 SUB Developer Kit},
  year         = {2025},
  url          = {https://category.yahboom.net/products/jetson-nano-sub},
  note         = {Accessed: 2025-12-07}
}

@misc{MacRumors_M2_2022,
  author       = {{MacRumors}},
  title        = {{M2} Chip Guide},
  year         = {2022},
  url          = {https://www.macrumors.com/guide/m2/},
  note         = {Accessed: 2025-06-12}
}

@InProceedings{iwann23,
author="Gamage, Udayanga K. N. G. W.
and Zanatta, Luca
and Fumagalli, Matteo
and Cadena, Cesar
and Tolu, Silvia",
editor="Rojas, Ignacio
and Joya, Gonzalo
and Catala, Andreu",
title="Event-Based Classification of Defects in Civil Infrastructures with Artificial and Spiking Neural Networks",
booktitle="Advances in Computational Intelligence",
year="2023",
publisher="Springer Nature Switzerland",
address="Cham",
pages="629--640",
isbn="978-3-031-43078-7"
}

@article{Gamage2025_evCIVIL,
  author  = {Gamage, Udayanga G. W. K. N. and Huo, Xuanni and Zanatta, Luca and Delbr{\"u}ck, T. and Cadena, Cesar and Fumagalli, Matteo and Tolu, Silvia},
  title   = {Event-based Civil Infrastructure Visual Defect Detection: {ev-CIVIL} Dataset and Benchmark},
  journal = {arXiv preprint arXiv:2504.05679},
  year    = {2025}
}

@misc{MIT_NAS_Lecture07,
  author       = {{MIT HAN Lab}},
  title        = {Lecture 07 -- {Neural Architecture Search (Part I)} | {MIT 6.S965}},
  howpublished = {YouTube video},
  year         = {2020},
  url          = {https://www.youtube.com/watch?v=NQj5TkqX48Q}
}

@inproceedings{neuromporphic_constraints,
  author    = {Balaji, Adarsha and Das, Anup},
  title     = {Compiling spiking neural networks to mitigate neuromorphic hardware constraints},
  booktitle = {Proceedings of the IEEE/ACM International Conference on Computer-Aided Design (ICCAD)},
  year      = {2020},
  doi       = {10.1145/3400302.3415694}
}

@article{Barchid2022,
  author  = {Barchid, S. and Mennesson, J. and Eshraghian, J. and Dj{\'e}raba, C. and Bennamoun, M.},
  title   = {Spiking neural networks for frame-based and event-based single object localization},
  journal = {Neurocomputing},
  volume  = {559},
  pages   = {126805},
  year    = {2023},
  doi     = {10.1016/j.neucom.2023.126805}
}

@article{VicenteSola2025,
  author  = {Vicente-Sola, A. and Manna, D. L. and Kirkland, P. and Di Caterina, G. and Bihl, T. J.},
  title   = {Spiking Neural Networks for event-based action recognition: A new task to understand their advantage},
  journal = {Neurocomputing},
  volume  = {611},
  pages   = {128657},
  year    = {2025},
  doi     = {10.1016/j.neucom.2024.128657}
}

@inproceedings{LaneSNN,
  author    = {Viale, Alberto and Marchisio, Alberto and Martina, Maurizio and Masera, Guido and Shafique, Muhammad},
  title     = {{LaneSNNs}: Spiking Neural Networks for Lane Detection on the {Loihi} Neuromorphic Processor},
  booktitle = {Proceedings of the 2022 IEEE/RSJ International Conference on Intelligent Robots and Systems (IROS)},
  year      = {2022},
  pages     = {79--86},
  address   = {Kyoto, Japan},
  doi       = {10.1109/IROS47612.2022.9981034}
}

@inproceedings{coco,
  author    = {Lin, Tsung-Yi and Maire, Michael and Belongie, Serge and Hays, James and Perona, Pietro and Ramanan, Deva and Doll{\'a}r, Piotr and Zitnick, C. Lawrence},
  title     = {Microsoft {COCO}: Common Objects in Context},
  booktitle = {Computer Vision -- {ECCV} 2014},
  year      = {2014},
  pages     = {740--755},
  publisher = {Springer International Publishing},
  address   = {Cham, Switzerland},
  doi       = {10.1007/978-3-319-10602-1\_48}
}

@inproceedings{li2021dfl,
  author    = {Li, Xiang and Wang, Wenhai and Wu, Lijun and Chen, Shuo and Hu, Xiaolin and Li, Jun and Tang, Jinhui and Yang, Jian},
  title     = {Generalized Focal Loss: Learning Qualified and Distributed Bounding Boxes for Dense Object Detection},
  booktitle = {Proceedings of the IEEE/CVF Conference on Computer Vision and Pattern Recognition (CVPR)},
  year      = {2020},
  pages     = {12839--12848}
}

@misc{nvidia_jetson_orin_nano_super,
  author       = {{NVIDIA Corporation}},
  title        = {{Jetson Orin Nano Super Developer Kit}},
  year         = {2024},
  howpublished = {\url{https://www.nvidia.com/en-us/autonomous-machines/embedded-systems/jetson-orin/nano-super-developer-kit/}},
  note         = {Accessed: 2026-03-29}
}

@misc{halapoint_2024,
  author       = {Dan Cooney},
  title        = {Intel unveils 1.15bn neuron neuromorphic system {Hala Point}},
  journal      = {Data Center Dynamics},
  year         = {2024},
  month        = {Apr},
  day          = {17},
  howpublished = {\url{https://www.datacenterdynamics.com/en/news/intel-unveils-115bn-neuron-neuromorphic-system-hala-point/}},
  note         = {Accessed: 2026-03-29}
}

\clearpage

\appendix

\section{Qualititative Visualization of Detection results}
\renewcommand{\thefigure}{A.\arabic{figure}}
\setcounter{figure}{0}
\setcounter{table}{0}
\renewcommand{\thetable}{A.\arabic{table}}
\setcounter{equation}{0}
\renewcommand{\theequation}{A.\arabic{equation}}
\label{appendix:a}

\begin{figure*}[!t]
\centering
\includegraphics[width=1.0\textwidth]{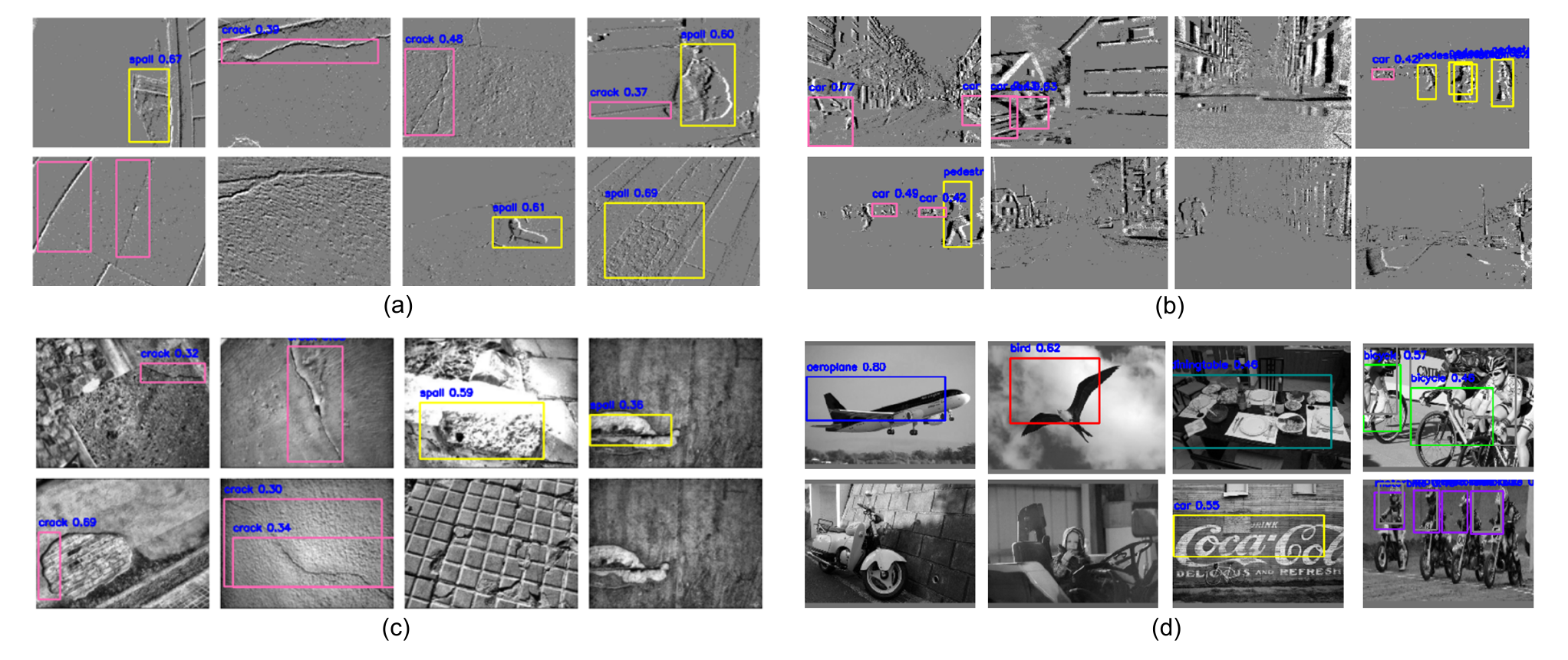}%
\caption{Qualitative visualization of sample detections on (a) evCIVIL-ev, (b) GEN1, (c) evCIVIL-fr, and (d) PASCAL VOC datasets. For each dataset, the first row shows the correct detections, while the second row illustrates the errors or misdetections. Different colored bounding boxes indicate detections of different object classes.}

\label{fig:qualititative_visualize_4datasets}
\end{figure*}

In \cref{fig:qualititative_visualize_4datasets}, for each of the dataset (evCIVIL-ev,evCIVIL-fr, PASCAL VOC and GEN1),  first row of images shows correct detections while the second row shows detection issues such as error detections and misdetections. 
In the PASCAL VOC dataset, the model exhibits detection issues, such as false detections such as detecting a Coca-Cola wall painting as a car, producing duplicate detections (e.g., in cyclist images), and misdetections of objects like scooters. 
In GEN1, misdetections of cars and humans are more common than false positives. 
For the evCIVIL-fr dataset, the model shows misclassification between crack and spalling defects, along with occasional duplicate crack detections. In the evCIVIL-ev dataset,small crack clusters are sometimes incorrectly identified as spalling, and additional misdetections observed between crack and spalling defects.

\section{Loss function definitions}

Here we explain the loss functions used in YOLOv8 related to the detection task.

\renewcommand{\thefigure}{B.\arabic{figure}}
\setcounter{figure}{0}
\setcounter{table}{0}
\renewcommand{\thetable}{B.\arabic{table}}
\setcounter{equation}{0}
\renewcommand{\theequation}{B.\arabic{equation}}
\label{appendix:d}

\subsection*{Box Regression Loss}

The bounding box regression loss($L_{box}$) which measures the geometric misalignment between predicted and target boxes and defined as in the \cref{eq:yolov8_box}.

\begin{equation}
\label{eq:yolov8_box}
\mathcal{L}_{\text{box}} = \frac{1}{S} \sum_{i \in \text{FG}} w_i \left( 1 - \text{CIoU}(B_i^p, B_i^t) \right)
\end{equation}

where:

\begin{itemize}
    \item $CIOU$ denote the Complete Intersection-over-Union based loss define in [8]
    \item $B_i^p$ and $B_i^t$ denote the predicted and target bounding boxes for sample $i$,
    \item $w_{i}$ is the confidence-based weight derived from the TTA (Task-Aligned Assigner) at foreground sample i.
    \item $S = \sum_{i \in \text{FG}} w_i$ is a normalization term over all foreground samples (FG).
\end{itemize}

\subsection*{Distribution Focal Loss(DFL)}

In anchor-free detection heads, which typically predict bounding box coordinates directly from feature map points without relying on predefined anchors, the Distribution Focal Loss ($L_{dfl}$) which is defined in \cref{eq:yolov8_dfl} is used to improve bounding box regression.
Here, instead of directly regressing continuous coordinates, DFL discretizes each bounding box coordinate into K bins and learns a probability distribution over these bins.
So for a given batch the DFL loss is defined as follows.

\begin{equation}
\label{eq:yolov8_dfl}
L_{\text{dfl}} = \frac{1}{S} \sum_{i \in \text{FG}} w_i \cdot \sum_{k \in \{l, r, t, b\}} \text{LDFL}\big(P_k(i), y_k(i)\big)
\end{equation}

\noindent
\textbf{where:}
  $S = \sum_{i \in \text{FG}} w_i$ is a normalization term over all foreground samples (FG).
  \( i \in \text{FG} \): iterates over all positive (foreground) anchor points.
  \( w_i \): the confidence-based weight for sample \( i \) (approximately the IoU of the predicted box).
   \( k \in \{l, r, t, b\} \): iterates over the four bounding box coordinates (left, right, top, bottom).
  \( \text{LDFL}(P_k(i), y_k(i)) \): the Distribution Focal Loss for a single coordinate as define in [8]

\subsection*{Binary Cross Entrophy With Logit Loss}

The classification loss in YOLOv8 is formulated using the Binary Cross-Entropy with Logits Loss ($L_{CLS}$) is defined as in \cref{eq:yolov8_cls}.

\begin{equation}
\label{eq:yolov8_cls}
L_{\text{cls}} = \sum_{i \in B} \sum_{c=1}^{C} L_{i,c}
\end{equation}

\begin{equation}
L_{i,c} = - \big[ y_{i,c} \cdot \log\big(\sigma(x_{i,c})\big) + (1 - y_{i,c}) \cdot \log\big(1 - \sigma(x_{i,c})\big) \big]
\end{equation}

\noindent
\textbf{where:}
\begin{itemize}
  \item \( L_{\text{CLS}} \): total classification loss.
  \item \( i \in B \): iterates over all samples in the batch.
  \item \( c \in \{1, \dots, C\} \): iterates over all classes.
  \item \( L_{i,c} \): binary cross-entropy loss for sample \( i \) and class \( c \).
  \item \( y_{i,c} \in \{0, 1\} \): ground truth label (1 if the class is present, otherwise 0).
  \item \( x_{i,c} \): raw (unnormalized) logit output of the model for class \( c \).
  \item \( \sigma(x) = \frac{1}{1 + e^{-x}} \): the sigmoid activation function.
\end{itemize}


\subsection*{KL divergence loss}

 we provide the definitions of the softmax probability distribution and the Kullback–Leibler (KL) divergence in \cref{eq:softmax} and \cref{eq:L_KL} respectively, based on which the distillation losses are defined.

Given the raw logits $Z_e \in \mathbb{R}^{M \times K}$, the softmax probability distribution $p_e$ is defined as:

\begin{equation}
p_e[i,j] = \frac{\exp(Z_e[i,j]/Tp)}{\sum_{k=1}^{K} \exp(Z_t[i,k]/Tp)}
\label{eq:softmax}
\end{equation}

where:\\
\(i = 1, \dots, M\), \(j = 1, \dots, K\) with M data samples and K channels\\
\(\,p_e[i,j]\) = probability of the \(i\)-th data sample belonging to class \(j\), \\
\(Z_e[i,j]\) = raw score (logit) for the \(i\)-th data sample and \(j\)-th class, \\
\(Tp\) = temperature parameter controlling the sharpness of the distribution.

where $\odot$ denotes the elementwise multiplication and $K$ is the size of the second dimension of $P_t$ and $P_s$ as introduced in \cref{eq:softmax}.

Given the ANN teacher and SNN student softmax distributions $p_t, p_s \in \mathbb{R}^{M \times K}$ defined in \cref{eq:softmax}, the KL divergence between the student distribution (SNN's predicted logits) $P_s$ and teacher distribution (ANN's predicted logits) $P_t$ over all data points can be expressed as:

\begin{equation}
D_{\mathrm{KL}}(P_s \,\|\, P_t) = \frac{1}{M} \sum_{i=1}^{M} \sum_{j=1}^{K} p_t[i,j] \, \log \frac{p_t[i,j]}{p_s[i,j]}
\label{eq:L_KL}
\end{equation}

where: 
\begin{itemize}
    \item $i = 1, \dots, M$ = $X \times Y$ where $X$ and $Y$ spatial dimensions of the output logit feature maps
    \item $j = 1, \dots, K$ denotes the number of feature maps/channels
    \item $p_t[i,j]$ is the probability of the $i$-th data sample belonging to class $j$ predicted by the teacher as defined in \cref{eq:softmax},
    \item $p_s[i,j]$ is the probability predicted by the student (SNN in our case) as defined in \cref{eq:softmax}.
\end{itemize}
\end{document}